\documentclass[11pt]{onesearchvl}

\usepackage{etoc}
\usepackage{tabularx}
\usepackage{array}
\usepackage{xurl}

\ifPDFTeX\else
  \newfontfamily\OneSearchKeepMedium[Path=./]{ByteSans-Medium.ttf}
  \newfontfamily\OneSearchKeepBold[Path=./]{ByteSans-Bold.ttf}
\fi
\newsavebox{\OneSearchAssetBox}
\AtBeginDocument{%
  \sbox{\OneSearchAssetBox}{%
    \includegraphics[height=1pt]{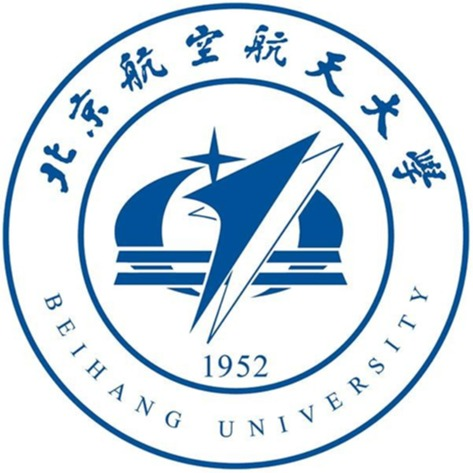}%
    \includegraphics[height=1pt]{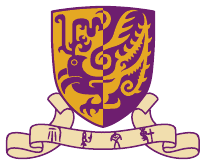}%
    \includegraphics[height=1pt]{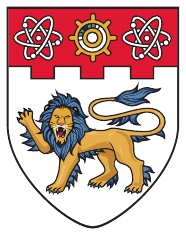}%
    \includegraphics[height=1pt]{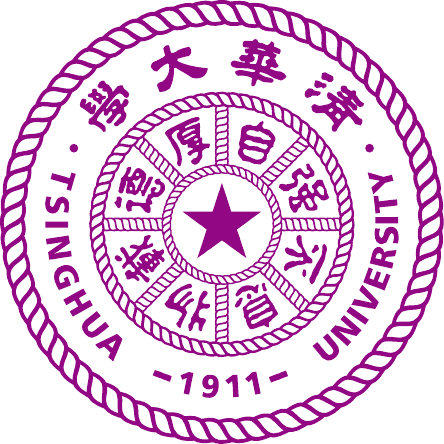}%
    \includegraphics[height=1pt]{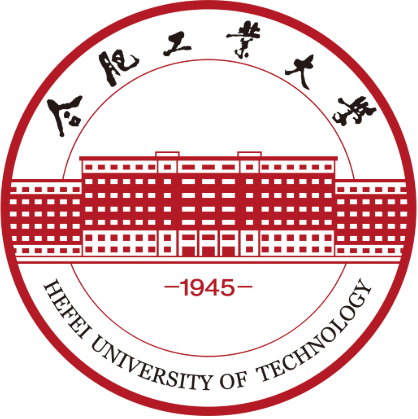}%
  }%
}

\title{\gradtext{OneSearch-VL}: Unified Multimodal Deep\\
Research Agent for Image and Video}

\author[1]{Hongyu Li}
\author[\dagger]{Manyuan Zhang}
\author[2]{Kaituo Feng}
\author[2]{Shu Chen}
\author[2]{Dian Zheng}
\author[3]{\\Hao Li}
\author[4]{Hao Yu}
\author[4]{Zhangquan Chen}
\author[2]{Zoey Guo}
\author[2]{Ray Zhang}
\author[5]{\\Shaofei Huang}
\author[5]{Tianrui Hui}
\author[1]{Linjiang Huang}
\author[1,\ddagger]{Si Liu}

\affiliation[1]{BUAA}
\affiliation[2]{CUHK}
\affiliation[3]{NTU}
\affiliation[4]{THU}
\affiliation[5]{HFUT}
\contribution[\dagger]{Project leader}
\contribution[\ddagger]{Corresponding author}

\abstract{Single-image, multi-image, and video deep research require different visual operations but share a workflow of visual grounding, external retrieval, and fact composition. A key challenge is to preserve the dependencies linking localized visual anchors, entity relations, source-supported facts, and answer-producing operations. We introduce \textbf{OneSearch-VL}, a unified agent centered on the Visually Grounded Evidence Graph (VGEG), which encodes these dependencies as a shared task-level reference for data construction, process supervision, and operation-level evaluation. Our VGEG-based data engine constructs and verifies multi-image and video questions and filters expert trajectories. Using these data, we assemble \textit{OneSearch-VL-SFT-110K} and \textit{OneSearch-VL-RL-10K} for SFT and RL, respectively. We further derive the Evidence-aware Visual-Grounded Rubric reward (EVGR) from VGEG annotations to supervise evidence traceability and visual grounding during RL. For finegrained evaluation, we construct \textit{OneSearch-MI-Bench} and \textit{OneSearch-Video-Bench}, organizing questions by the research operations encoded in their VGEGs. Experiments show that OneSearch-VL-8B improves over Qwen3-VL-8B with tool access by 20.2 and 17.6 percentage points on the two new benchmarks, respectively, while also achieving substantial gains across 7 image benchmarks and VideoDR.
}

\colorlet{OneSearchLinkColor}{onesearchpurple}
\checkdata[\raisebox{-1.8pt}{\includegraphics[height=10pt]{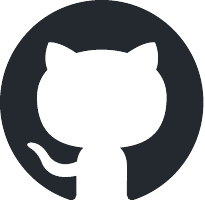}}\hspace{0.35em}Home]{\textcolor{OneSearchLinkColor}{\url{https://github.com/appletea233/OneSearch-VL}}}
\checkdata[\raisebox{-1.8pt}{\includegraphics[height=10pt]{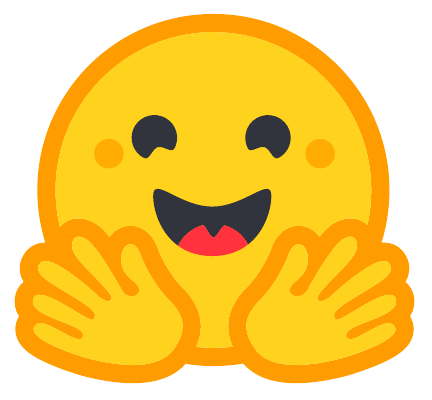}}\hspace{0.35em}Hugging Face]{\textcolor{OneSearchLinkColor}{\url{https://huggingface.co/OneSearch-VL}}}

\hypersetup{
  pdftitle={OneSearch-VL: Unified Multimodal Deep Research Agent for Image and Video},
  pdfauthor={Hongyu Li, Manyuan Zhang, Kaituo Feng, Shu Chen, Dian Zheng, Hao Li, Hao Yu, Zhangquan Chen, Zoey Guo, Ray Zhang, Shaofei Huang, Tianrui Hui, Linjiang Huang, Si Liu}
}

\begin{document}
\etocdepthtag.toc{main}
\maketitle

\section{Introduction}

Multimodal deep research enables vision-language models to move beyond direct answering by actively locating visual clues, retrieving external knowledge, and integrating evidence through multi-turn tool use \citep{wu2025mmsearchr1,geng2025webwatcher,zhou2026visiondeepresearch,chen2026opensearchvl}. Recent work has extended this capability from images to videos by combining temporal localization, visual inspection, and open-web retrieval \citep{liu2026videodr,gao2026videosearcher,fang2026videodeepresearch}. Meanwhile, unified multimodal models extend spatial-temporal understanding and reasoning across images and videos \citep{li2025llavast,feng2025onethinker}. These advances raise a natural question: can one policy jointly learn deep research over both images and videos? Although these settings require different visual operations, they share a workflow from visual grounding to external search and fact composition, motivating unified training across multiple visual input types.

Moving beyond prior work that primarily considers single-image or video inputs, we additionally study multi-image deep research and develop a unified framework spanning all three visual input types. Training data for this unified setting must link visual entities distributed across regions, images, or frames with relevant facts scattered across webpages. Prior work represents evidence through perception--knowledge chains, factual rubrics, and evidence graphs for task construction, supervision, and evaluation \citep{searches2026searchworld,zhang2026carr,xu2026stamp,sun2026hievidr}. However, these structures do not jointly preserve the dependencies from localized visual anchors, through entity relations and source-supported facts, to answer composition. Video-centric pipelines similarly use extracted entities or key-frame crops mainly as retrieval or question-generation seeds \citep{gao2026videosearcher,fang2026videodeepresearch}, without explicitly retaining this end-to-end provenance.

To address this gap, we introduce the \textbf{Visually Grounded Evidence Graph (VGEG)}, a unified task-level reference structure for single-image, multi-image, and video research. VGEG links localized visual anchors to real-world entities, records multi-hop relational paths and source-supported facts, and specifies the operations that compose these facts into an answer. Based on VGEG, we develop a data engine for single-image, multi-image and video research that extracts visual entities, retrieves and verifies web facts, generates compositional questions, and synthesizes expert tool-use trajectories while retaining visual locations, source provenance, and evidence dependencies. These retained dependencies support question-answer construction and verification, expert-trajectory filtering, fine-grained trajectory rewards for RL, and operation-level evaluation. With this data engine, we construct \textbf{OneSearch-VL-SFT-110K} and \textbf{OneSearch-VL-RL-10K} for SFT and RL.

Using these data, we develop \textbf{OneSearch-VL}, a unified agent trained on data spanning single-image, multi-image, and video inputs through SFT and RL. Existing multimodal agents supervise search through outcome rewards, query-quality signals, tool-use objectives, intermediate entity anchors, or evidence rubrics. We further introduce the \textbf{Evidence-aware Visual-Grounded Rubric reward (EVGR)}, constructed from our VGEG annotations to evaluate complete trajectories along two complementary dimensions: \emph{evidence traceability}, which verifies whether the required facts are supported by tool observations, and \emph{visual grounding}, which verifies whether the relevant objects, regions, or frames are correctly identified and used in the solution. 

For fine-grained evaluation, we also introduce \textbf{OneSearch-MI-Bench} and \textbf{OneSearch-Video-Bench}. Existing video deep-research benchmarks primarily evaluate answer accuracy, with analyses commonly organized around video content or source categories \citep{liu2026videodr,gao2026videosearcher}. 
Our benchmarks instead organize questions according to the research operations represented in their corresponding VGEGs, which capture the entity-relation paths and fact-composition dependencies required to derive the answer. This organization emphasizes how an agent retrieves and composes evidence rather than what the visual input contains.

We evaluate OneSearch-VL on established and two new benchmarks. OneSearch-VL-8B outperforms Qwen3-VL-8B by 20.2, 17.6, and 27.0 points on OneSearch-MI-Bench, OneSearch-Video-Bench, and VideoDR, respectively. Joint SFT on all three visual input types achieves a six-benchmark average of 55.8, versus 51.7-55.1 for single-type training. EVGR further raises the average from 57.3 to 61.1 over answer and query rewards.

Our contributions are summarized as follows:
\begin{itemize}
    \item \textbf{VGEG-centered data engine.} We propose VGEG, a unified structure linking visual anchors, source-supported facts, and answer composition, together with a data engine for constructing and verifying multi-image and video questions and expert trajectories.
    \item \textbf{Unified agent and evidence supervision.} We develop OneSearch-VL for joint single-image, multi-image, and video deep research, and use EVGR to incorporate evidence traceability and visual grounding into RL.
    \item \textbf{Operation-oriented benchmark.} We construct OneSearch-MI-Bench and OneSearch-Video-Bench benchmarks spanning research operations for fine-grained evaluation of fact retrieval and composition.
\end{itemize}
\section{Unified Multimodal Deep Research}
\label{sec:unified_research}

\subsection{Problem Formulation}

We study multimodal deep research over three types of visual input. Given a visual input $X$ and a question $q$, an agent actively inspects the visual content, retrieves external information, and synthesizes a supported answer. The input is either a single image, an image collection, or a video:
\begin{equation}
X \in
\left\{
I,\;
\mathcal I=\{I_j\}_{j=1}^{K},\;
V
\right\}.
\end{equation}
We learn a unified policy $\pi_\theta$ that selects visual inspection, temporal localization, and web retrieval operations according to the input and question.

At step $t$, the policy conditions on the interaction history
$h_t=(X,q,a_1,o_1,\ldots,a_{t-1},o_{t-1})$
and generates an action $a_t=(z_t,c_t)$, where $z_t$ denotes the reasoning trace and $c_t$ is either a tool call or the final response. A tool call produces an observation $o_t$, such as a cropped region, a video frame, recognized text, or retrieved information. The complete trajectory is
$\tau=(X,q,a_1,o_1,\ldots,a_{T-1},o_{T-1},a_T)$
where $a_T$ contains the final answer. Its likelihood is defined as:
\begin{equation}
\pi_\theta(\tau\mid X,q)
=
\prod_{t=1}^{T}\pi_\theta(a_t\mid h_t).
\end{equation}

All three types of visual input share the same action--observation protocol while retaining input-specific behaviors: single-image tasks inspect local regions, multi-image tasks aggregate information across indexed images, and video tasks localize relevant temporal segments and frames.

\subsection{Tool Environment}

OneSearch-VL exposes tools according to the visual input type, as detailed in Table~\ref{tab:tool-environment}:
\begin{equation}
\mathcal T(X)=
\begin{cases}
\mathcal T_{\mathrm{vis}}\cup\mathcal T_{\mathrm{ret}},
& X\in\{I,\mathcal I\},\\
\mathcal T_{\mathrm{vis}}\cup\mathcal T_{\mathrm{ret}}
\cup\mathcal T_{\mathrm{temp}},
& X=V.
\end{cases}
\end{equation}

All tools share a common dispatcher and observation format. This interface allows the policy to compose operations specific to each visual input type while using the same trajectory representation for expert synthesis, reinforcement learning, and inference. Detailed execution semantics are provided in Appendix~\ref{app:tool-details}.

\begin{table}[t]
\vspace{-1em}
\centering
\caption{\textbf{Tool suite of OneSearch-VL.}}
\label{tab:tool-environment}

\footnotesize
\setlength{\tabcolsep}{3pt}
\renewcommand{\arraystretch}{1.08}

\resizebox{\linewidth}{!}{%
\begin{tabular}{@{}llllc@{}}
\toprule
\textbf{Tool Set}
&
\textbf{Tool}
&
\textbf{Description}
&
\textbf{Arguments}
&
\textbf{Visual Input Type}
\\
\midrule

\multirow{5}{*}{$\mathcal T_{\mathrm{vis}}$}
& \textsc{Crop}
& Crop an image region.
& Image + Coordinates
& \multirow{5}{*}{$I,\mathcal I,V$} \\

& \textsc{OCR}
& Extract visible text and layout.
& Image
& \\

& \textsc{PerspectiveCorrect}
& Correct perspective distortion.
& Image
& \\

& \textsc{SuperResolution}
& Upscale low-resolution images.
& Image + Scale
& \\

& \textsc{Sharpen}
& Reduce blur and enhance details.
& Image + Amount
& \\

\addlinespace[3pt]

\multirow{2}{*}{$\mathcal T_{\mathrm{ret}}$}
& \textsc{ImageSearch}
& Identify entities through image search.
& Image
& \multirow{2}{*}{$I,\mathcal I,V$} \\

& \textsc{TextSearch}
& Search, read, and summarize webpages.
& Query + Language + TopK
& \\

\addlinespace[3pt]

\multirow{2}{*}{$\mathcal T_{\mathrm{temp}}$}
& \textsc{SelectTimespan}
& Select a contiguous video interval.
& Video + Start + End
& \multirow{2}{*}{$V$} \\

& \textsc{SelectFrame}
& Extract a frame by index.
& Video + Frame Index
& \\

\bottomrule
\end{tabular}%
}
\vspace{-1.5em}
\end{table}

\section{VGEG-Centered Multimodal Data Engine}
\label{sec:data-engine}

\begin{figure}[!t]
    \centering
    \includegraphics[width=0.85\linewidth]{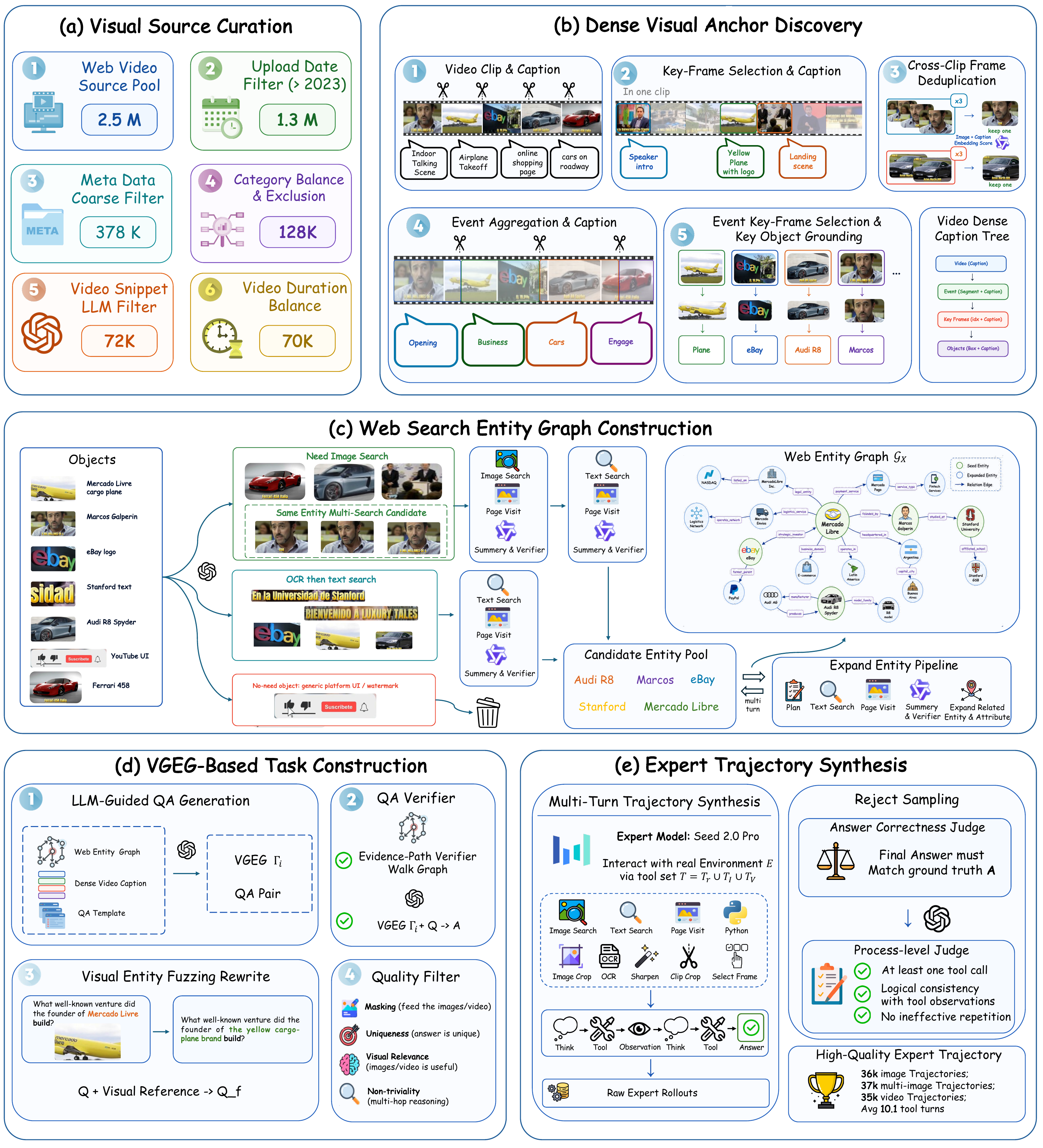}
    \caption{\textbf{Overview of the VGEG-centered multimodal data engine.} (a) Visual source curation retains videos suitable for external-knowledge research. (b) Dense visual anchor discovery organizes each video into events, key frames, and localized objects. (c) Web entity graph construction links these anchors to real-world entities, source-supported facts, and webpages, forming the evidence graph $\mathcal G_X$. (d) VGEG-based task construction produces VGEG $\Gamma_i$ for question generation, evidence verification, and visual-reference rewriting. (e) Expert trajectories in the tool environment are filtered by answer correctness and process quality to yield multi-turn training trajectories.}
    \label{fig:data-pipeline}
\end{figure}

Figure~\ref{fig:data-pipeline} presents the complete data engine. Starting from raw visual sources, the pipeline identifies localized visual anchors, connects them to source-supported web facts, constructs task-level VGEGs, and synthesizes verified question--answer pairs and expert trajectories. See Appendix~\ref{app:data-engine} for details.

\subsection{Visual Source Curation}

As shown in Figure~\ref{fig:data-pipeline}(a), we first curate source videos that contain identifiable entities, concrete event cues, and sufficient potential for external knowledge retrieval. The pipeline combines metadata screening, task-suitability assessment, category balancing, and duration balancing, reducing the initial pool of 2.5M web videos to 70k candidates. See Appendix~\ref{app:source-filtering} for details.

\subsection{Dense Visual Anchor Discovery}

Figure~\ref{fig:data-pipeline}(b) converts each retained video into a hierarchical visual representation. We sample and group frames into local clips, generate clip- and frame-level descriptions, remove redundant key frames, aggregate temporally related clips into events, and localize searchable objects in representative frames. The resulting dense caption tree links each event to its temporal range, key frames, and localized object instances, which serve as visual anchors for subsequent retrieval. See Appendix~\ref{app:visual-anchors} for details.

\subsection{Web Evidence Graph Construction}

As illustrated in Figure~\ref{fig:data-pipeline}(c), we connect visual anchors to external knowledge through image search, OCR, and text retrieval. Candidate identities are verified against the visual context and retrieved sources, after which source-supported relations and attributes are extracted and iteratively expanded. These results form an input-level evidence graph $\mathcal G_X$ that preserves the correspondence among visual anchors, real-world entities, external facts, and supporting sources. See Appendix~\ref{app:web-evidence} for details.

\subsection{VGEG-Based Task Construction}

Figure~\ref{fig:data-pipeline}(d) constructs candidate research tasks and their task-level references from the input-level graph $\mathcal G_X$. For each candidate task $i$, its Visually Grounded Evidence Graph (VGEG) is derived through a task-conditioned projection $\Phi_i$ that selects the required visual anchors and facts from $\mathcal G_X$ and augments them with explicit answer-producing operations:
\begin{equation}
    \Gamma_i = \Phi_i(\mathcal G_X)
    =
    \left(
    \mathcal A_i,
    \mathcal F_i,
    \mathcal O_i,
    \mathcal R_i
    \right).
\end{equation}
Here, $\mathcal A_i$ contains localized visual anchors and their associated entities, $\mathcal F_i$ contains source-supported external facts, $\mathcal O_i$ specifies how the selected facts produce the answer, and $\mathcal R_i$ combines grounding and support relations inherited from $\mathcal G_X$ with task-specific dependencies among anchors, facts, and operations.

The generator jointly outputs $(q_i,y_i,\Gamma_i)$ from $\mathcal G_X$. We verify the question and answer against the selected evidence and rewrite explicit entity mentions into visually grounded references. We then instantiate each task for its target visual input type. For a multi-image task, the deduplicated key frames containing the required anchors form an unordered image collection, and frame-level references are remapped to image indices. For a video task, the original video is retained together with its event, timestamp, frame, and region bindings. Additional filters remove answer leakage, ambiguous references, visually irrelevant questions, and instances that can be confidently answered without the visual input. See Appendix~\ref{app:vgeg-construction} for details. For single-image task, we reuse existing data~\citep{chen2026opensearchvl} and convert its Wikipedia entity-sampling paths and associated visual anchors into  VGEGs.

\subsection{Expert Trajectory Synthesis}

Finally, as shown in Figure~\ref{fig:data-pipeline}(e), we synthesize expert trajectories for the verified tasks. An expert model receives only the visual input, question, and available tools and interacts with the environment until producing a final answer or reaching the interaction budget. We retain trajectories that pass both answer-correctness and process-quality evaluation and combine them with single-image trajectories in a unified multi-turn format to construct \textbf{OneSearch-VL-SFT-110K}. See Appendix~\ref{app:trajectory-synthesis} for details. For RL, we construct \textbf{OneSearch-VL-RL-10K} from task pools spanning the same three visual input types, with construction details provided in Appendix~\ref{app:training-details}.
\section{Onesearch-Bench}
\label{sec:benchmarks}

\begin{figure*}[t]
    \centering
    \includegraphics[width=\textwidth]{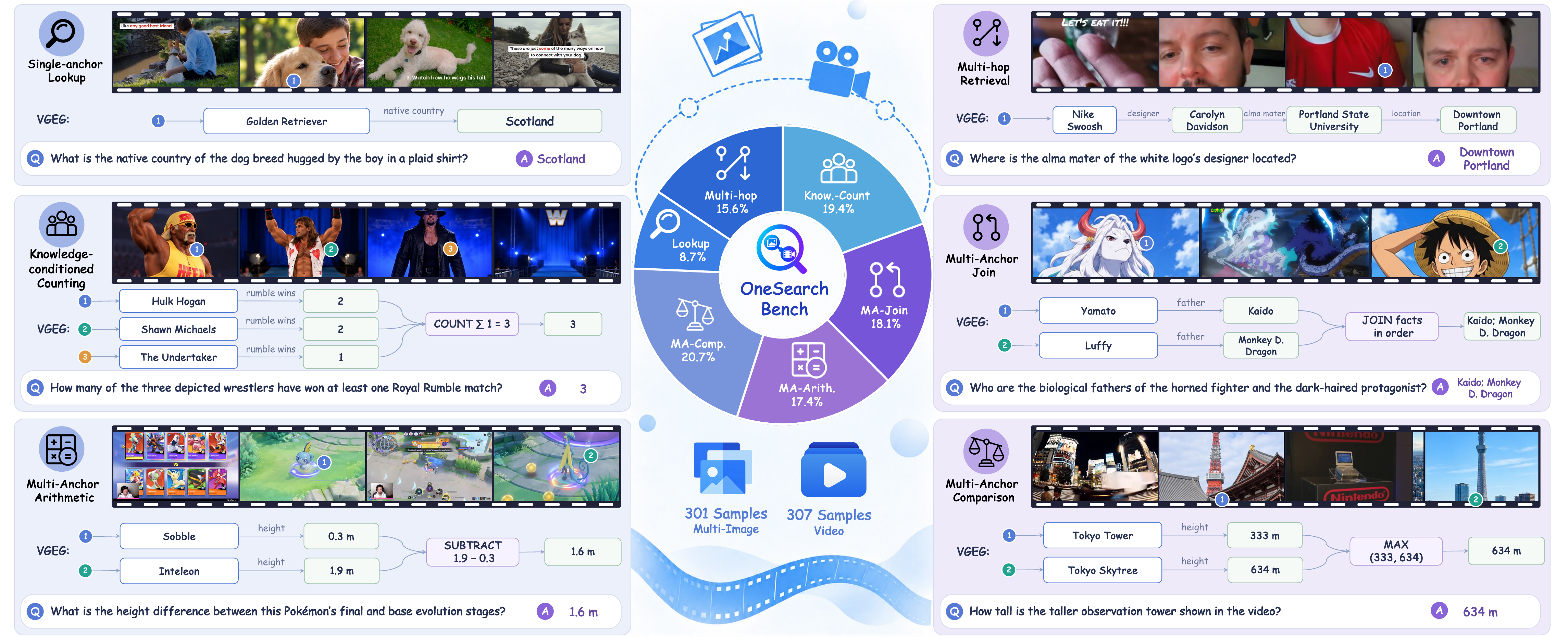}
    \caption{\textbf{Operation-oriented design of the OneSearch benchmarks.} Representative questions and reference VGEGs are shown for six principal research operations. Numbered markers bind visual evidence to graph nodes; green nodes denote retrieved entity/fact and purple nodes denote answer-producing operations. The center summarizes the operation distribution and the sizes of the multi-image and video benchmarks.}
    \label{fig:benchmark-overview}
\end{figure*}

Existing multimodal search and video deep-research benchmarks primarily report aggregate answer accuracy, often with additional breakdowns by visual content or data source \citep{liu2026videodr,gao2026videosearcher}. These summaries measure overall performance but do not isolate failures in entity lookup, relation tracing, conditional filtering, or fact composition. We therefore construct \textbf{OneSearch-MI-Bench} and \textbf{OneSearch-Video-Bench} and organize their examples by the \emph{principal research operation} required to answer each question. Both benchmarks use the same operation taxonomy, enabling a consistent analysis across multi-image and video inputs.

Figure~\ref{fig:benchmark-overview} presents representative examples and their reference VGEGs, while Appendix Table~\ref{tab:benchmark-structure} defines the six principal research operations and reports their distributions across the two benchmarks. Together, they support operation-level analysis of evidence retrieval and composition.

For {OneSearch-MI-Bench}, deduplicated evidence frames from the same visual source form an unordered image set $X=\{I_1,\ldots,I_K\}$. Each example depends on at least two images and retains image indices that bind visual anchors to their sources. We remove references to video playback, frame numbers, and unnecessary temporal order so that each question is self-contained over the image set, while preserving the relations needed for fact composition.

For {OneSearch-Video-Bench}, we preserve temporal structure and link visual anchors to their events, key frames, and web facts. Questions locate clues through visible objects, events, or temporal descriptions, requiring the agent to find relevant video content before retrieving and composing external knowledge. The two branches share candidate sources, VGEG representations, the operation taxonomy, and quality criteria, while producing separate multi-image and video sets.

Candidates pass checks for information masking, referential uniqueness, visual relevance, and non-triviality. A text-only probe further removes questions that can be answered confidently without the visual input. We then jointly inspect the question, reference answer, and supporting facts for clarity, answer uniqueness, factual correctness, and visual grounding, followed by a strict review of the visual input.

\section{OneSearch-VL Training}
\label{sec:training}
\begin{figure}[t]
    \centering
    \includegraphics[width=0.99\linewidth]{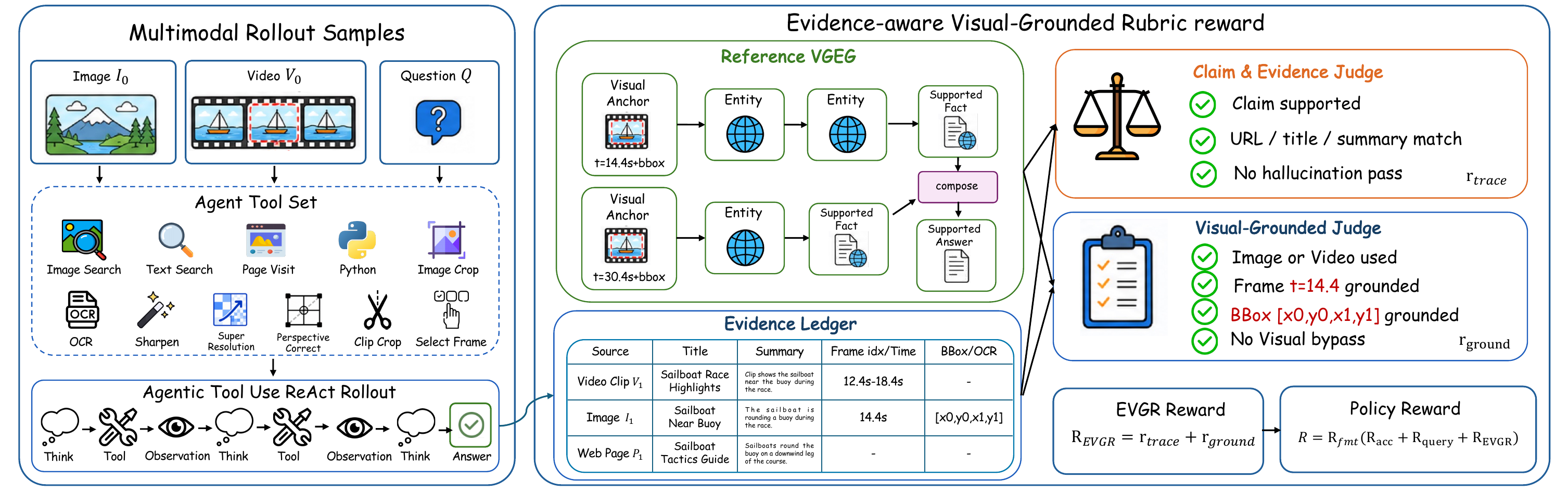}
    \caption{\textbf{Overview of the Evidence-aware Visual-Grounded Rubric reward (EVGR).} A task-level VGEG and evidence ledger align each multimodal rollout with the required visual anchors and source-supported facts. Claim-and-evidence and visual-grounding judges produce $r_{\mathrm{trace}}$ and $r_{\mathrm{ground}}$, whose combination forms $R_{\mathrm{EVGR}}$ for policy optimization.}
    \label{fig:evgr}
\end{figure}

We train OneSearch-VL in two stages. Supervised fine-tuning learns a unified multimodal tool-interaction policy from expert trajectories, and reinforcement learning further optimizes answer quality, search behavior, and evidence use through online interaction with tool environment.

\subsection{Agentic Supervised Fine-Tuning}

We first perform SFT on \textit{OneSearch-VL-SFT-110K}. Each trajectory interleaves reasoning, tool calls, tool observations, and a final answer. All three visual input types share the same action--observation schema, allowing one policy to select input-appropriate tools within a unified action space.

Tool observations are exogenous environment outputs and serve only as conditioning context. Let $y_{t,k}$ denote the $k$-th policy-generated token at turn $t$, and let $M_{t,k}^{\mathrm{pol}}$ be one for reasoning, tool-command, and final-response tokens and zero for serialized tool observations. The SFT objective is
\begin{equation}
    \mathcal{L}_{\mathrm{SFT}}(\theta)
    = -\sum_{\tau}\sum_t\sum_k
    M_{t,k}^{\mathrm{pol}}
    \log \pi_\theta(y_{t,k}\mid h_t,y_{t,<k}).
\end{equation}
This stage yields the initial policy $\pi_{\theta_{\mathrm{SFT}}}$ for reinforcement learning.

\subsection{Reinforcement Learning}

Starting from $\pi_{\theta_{\mathrm{SFT}}}$, we perform online rollouts on \textit{OneSearch-VL-RL-10K} with the same dispatcher and tool environment used for expert trajectory synthesis and inference. Given visual input $X$, question $Q$, and reference answer $A$, the behavior policy samples a group of trajectories
\begin{equation}
    \tau_i \sim
    \pi_{\theta_{\mathrm{old}}}(\cdot\mid X,Q;\mathcal{E}),
    \qquad i=1,\ldots,G.
\end{equation}

\paragraph{\normalfont\textbf{Dual-dimensional EVGR.}}

Terminal correctness does not directly measure whether a rollout obtains and uses the evidence required by the question. We therefore introduce the Evidence-aware Visual-Grounded Rubric reward (EVGR). As illustrated in Figure~\ref{fig:evgr}, EVGR aligns each executed trajectory with an item-specific structured rubric derived from the data engine. The rubric records the answer core, required claim facts, supporting sources and snippets, and applicable frame or region annotations. Two separate judge calls evaluate complementary dimensions. \textbf{Evidence Traceability} ($r_{\mathrm{trace}}$) measures whether tool observations establish the answer-core entities and required fact hops, and whether the reasoning follows those observations without unsupported substitutions. \textbf{Visual Grounding} ($r_{\mathrm{ground}}$) measures whether the correct visible entities, regions, or video frames are identified and used to drive subsequent evidence retrieval. For single- and multi-image inputs, the source images are already visible to the agent policy and no additional visual tool call is required; for video, the selected frames should cover the relevant moments and provide the visual identity used by later searches. The two dimensions are combined into the trajectory-level process reward $R_{\mathrm{EVGR}}$. Their scoring and combination coefficients are specified in the experimental setup.

\paragraph{\normalfont\textbf{Composite reward and policy optimization.}}

Following OpenSearch-VL~\citep{chen2026opensearchvl}, we adopt its answer-correctness reward $R_{\mathrm{acc}}$ and query-quality reward $R_{\mathrm{query}}$, and augment them with our EVGR process reward. The resulting reward is gated by format validity:
\begin{equation}
    R(\tau)=R_{\mathrm{fmt}}(\tau)\Bigl(
    \lambda_{\mathrm{acc}}R_{\mathrm{acc}}(\tau)
    +\lambda_{\mathrm{query}}R_{\mathrm{query}}(\tau)
    +\lambda_{\mathrm{EVGR}}R_{\mathrm{EVGR}}(\tau)
    \Bigr).
\end{equation}
Here, $R_{\mathrm{acc}}$ evaluates the final answer, $R_{\mathrm{query}}$ evaluates the relevance and progression of search queries, and $R_{\mathrm{fmt}}$ enforces a valid reasoning, tool-call, and response structure. The reward coefficients and implementation details are provided in Appendix~\ref{app:training-details}.

We optimize the policy with Group Relative Policy Optimization (GRPO), normalizing rewards across trajectories sampled for the same question and applying the clipped objective only to policy-generated tokens. Tool observations remain conditioning context and are excluded from the policy loss. For trajectories terminated by tool-execution failures, we follow the fatal-aware mechanism of OpenSearch-VL \citep{chen2026opensearchvl} and retain only the valid pre-failure prefix for optimization.

\section{Experiments}
\label{sec:experiments}

\subsection{Experimental Setup}

\paragraph{\normalfont\textbf{Model and training data.}}
OneSearch-VL is initialized from Qwen3-VL-8B~\citep{bai2025qwen3vl} and trained in two stages. Supervised fine-tuning uses \textit{OneSearch-VL-SFT-110K}. Reinforcement learning starts from the joint SFT model and uses \textit{OneSearch-VL-RL-10K}. Both datasets cover all three visual input types. All online rollouts use the same tool environment.

\paragraph{\normalfont\textbf{Benchmarks.}}
For single-image evaluation, we follow OpenSearch-VL~\citep{chen2026opensearchvl} and use seven knowledge-intensive benchmarks: SimpleVQA~\citep{cheng2025simplevqa}, VDR~\citep{vdr}, MMSearch~\citep{mmsearch}, LiveVQA~\citep{fu2025livevqa}, BrowseComp-VL~\citep{webwatcher}, FVQA~\citep{wang2017fvqa}, and InfoSeek~\citep{chen2023can}. For multi-image and video evaluation, we use our OneSearch-MI-Bench and OneSearch-Video-Bench, together with VideoDR~\citep{liu2026videodr}. On the two OneSearch benchmarks, we additionally report performance by the six principal research operations defined in Section~\ref{sec:benchmarks}. Following OpenSearch-VL, we use GPT-4o~\citep{openai2024gpt4ocard} to compare each final response with the reference answer and return a binary correctness decision.

\paragraph{\normalfont\textbf{Implementation details.}}
Expert trajectory synthesis, online RL rollouts, and inference share the same dispatcher and tool interfaces. All training variants are compared under the same tool environment and evaluation protocol. Full optimization, rollout, reward, and inference configurations are provided in the appendix.

\subsection{Main Results}

\paragraph{\normalfont\textbf{Single-image multimodal deep research.}}
Table~\ref{tab:image-main} summarizes the results on seven single-image deep-research benchmarks. OneSearch-VL-8B obtains an average score of 58.3, outperforming the same-scale Qwen3-VL-8B Agent by 16.3 points and OpenSearch-VL-8B by 1.7 points. It improves over OpenSearch-VL on all seven benchmarks, with gains of 2.8, 2.1, and 2.0 points on VDR, InfoSeek, and MMSearch, respectively. It also achieves the best result among the listed agentic methods on six of the seven benchmarks.

The gains span benchmarks emphasizing visual entity identification, open-web retrieval, and multi-step knowledge reasoning. Together with the data-mixture ablation below, these results show that joint training with multi-image and video trajectories does not trade off image research performance.

\begin{table}[t]
    \setlength{\abovecaptionskip}{0pt}
    \setlength{\belowcaptionskip}{0pt}
    \caption{{Results on single-image deep-research benchmarks.}  
    }
    \label{tab:image-main}
    \centering
    \footnotesize
    \setlength{\tabcolsep}{4.5pt}
    \renewcommand{\arraystretch}{1.08}
    \resizebox{0.98\linewidth}{!}{%
    \begin{tabular}{l ccccccc c}
    \toprule
    \textbf{Model} & \textbf{SimpleVQA} & \textbf{VDR} & \textbf{MMSearch} & \textbf{LiveVQA} & \textbf{BrowseComp-VL} & \textbf{FVQA} & \textbf{InfoSeek} & \textbf{Avg.} \\
    \midrule
    \rowcolor{blue!5}
    \multicolumn{9}{c}{\textbf{Direct Reasoning}} \\
    GPT-4o~\citep{openai2024gpt4ocard} & 51.7 & 1.7 & 18.7 & 28.1 & 5.5 & 48.0 & 52.9 & 29.5 \\
    GPT-5~\citep{openai2025gpt5} & \underline{61.6} & \textbf{9.8} & \underline{35.1} & 44.4 & \textbf{48.6} & \underline{54.4} & \textbf{61.7} & \underline{45.1} \\
    Gemini-2.5-Flash~\citep{comanici2025gemini} & 57.9 & 6.2 & 30.4 & \underline{51.0} & 37.1 & 47.7 & 44.1 & 39.2 \\
    Gemini-2.5-Pro~\citep{comanici2025gemini} & \textbf{63.0} & \underline{8.0} & \textbf{39.8} & \textbf{60.3} & \underline{43.1} & \textbf{60.7} & 46.9 & \textbf{46.0} \\
    Claude-4-Sonnet~\citep{anthropic2025claude4} & 50.9 & 2.0 & 18.7 & 38.5 & 29.3 & 35.3 & \underline{57.3} & 33.1 \\
    Claude-3.7-Sonnet~\citep{anthropic2025claude37} & 42.7 & 4.6 & 21.1 & 38.0 & 32.3 & 36.7 & 54.8 & 32.9 \\
    Qwen3-VL-8B~\citep{bai2025qwen3vl} & 47.1 & 2.8 & 11.7 & 23.1 & 24.1 & 24.2 & 23.1 & 22.3 \\
    Qwen3-VL-30B-A3B~\citep{bai2025qwen3vl} & 53.2 & 3.8 & 18.7 & 42.7 & 29.6 & 34.7 & 26.4 & 29.9 \\
    Qwen3-VL-32B~\citep{bai2025qwen3vl} & 58.0 & 4.1 & 19.8 & 45.5 & 30.8 & 34.1 & 28.8 & 31.6 \\
    \midrule
    \rowcolor{blue!5}
    \multicolumn{9}{c}{\textbf{RAG Workflow}} \\
    GPT-4o~\citep{openai2024gpt4ocard} & \textbf{63.6} & 4.5 & \underline{49.1} & \underline{40.1} & 13.4 & \textbf{66.3} & 59.5 & \underline{42.4} \\
    GPT-5~\citep{openai2025gpt5} & 55.9 & \textbf{22.3} & \textbf{52.6} & \textbf{56.0} & \textbf{54.9} & \underline{62.6} & \textbf{70.6} & \textbf{53.6} \\
    Claude-3.7-Sonnet~\citep{anthropic2025claude37} & 59.3 & \underline{11.3} & 32.7 & 30.3 & 10.0 & 59.1 & \underline{60.2} & 37.6 \\
    Qwen3-VL-8B~\citep{bai2025qwen3vl} & \underline{62.3} & 7.3 & 47.3 & 39.3 & \underline{29.3} & 53.6 & 46.1 & 40.7 \\
    \midrule
    \rowcolor{blue!5}
    \multicolumn{9}{c}{\textbf{Agentic Workflow}} \\
    DeepMMSearch-R1-7B~\citep{narayan2025deepmmsearchr1} & 55.8 & -- & -- & -- & -- & -- & 47.5 & -- \\
    Visual-ARFT-7B~\citep{liu2025visualarft} & 42.4 & 3.3 & 34.5 & 25.4 & 16.5 & 41.7 & 37.9 & 28.8 \\
    MMSearch-R1-7B~\citep{wu2025mmsearchr1} & 57.4 & 2.9 & 53.8 & 48.4 & 20.9 & 58.4 & 55.1 & 42.4 \\
    DeepEyes-v2-7B~\citep{hong2025deepeyesv2} & 59.4 & 7.8 & 63.7 & -- & -- & 60.6 & 51.1 & -- \\
    WebWatcher-7B~\citep{geng2025webwatcher} & 54.3 & 10.3 & 49.1 & 51.2 & 21.2 & -- & -- & -- \\
    Qwen3-VL-8B~\citep{bai2025qwen3vl} & 52.0 & 17.0 & 37.4 & 50.6 & 27.9 & 58.7 & 50.3 & 42.0 \\
    SenseNova-MARS-8B~\citep{chng2025sensenova} & 61.7 & 19.4 & \textbf{67.4} & 56.2 & 35.1 & 67.1 & 61.7 & 52.7 \\
    {OpenSearch-VL-8B}~\citep{chen2026opensearchvl} & \underline{71.6} & \underline{20.8} & 64.5 & \underline{59.6} & \underline{37.6} & \underline{71.5} & \underline{70.2} & \underline{56.6} \\
    \rowcolor{purple!12}
    \textbf{OneSearch-VL-8B (Ours)} & \textbf{72.4} & \textbf{23.6} & \underline{66.5} & \textbf{61.2} & \textbf{39.0} & \textbf{73.4} & \textbf{72.3} & \textbf{58.3} \\
    \bottomrule
    \end{tabular}%
    }
\end{table}

\paragraph{\normalfont\textbf{Multi-image and video deep research.}}
As shown in Table~\ref{tab:multimodal-bench-main}, OneSearch-VL-8B improves over Qwen3-VL-8B by 20.2 points on OneSearch-MI-Bench, 17.6 points on OneSearch-Video-Bench, and 27.0 points on VideoDR. It improves all six research operations on both OneSearch benchmarks, showing that the gains extend beyond simple fact lookup to questions that combine evidence across images or video frames.

The operation-level breakdown further identifies where the gains arise. On multi-image tasks, knowledge-conditioned counting, multi-anchor arithmetic, and multi-anchor comparison improve by 29.8, 26.3, and 22.5 points, respectively. On video tasks, multi-hop retrieval, multi-anchor arithmetic, and multi-anchor joining improve by 25.5, 20.0, and 19.2 points. These categories require the model to filter, connect, or compute over facts associated with multiple visual anchors, matching the compositional operations covered by our data engine.

\begin{table}[t]
    \setlength{\abovecaptionskip}{0pt}
    \setlength{\belowcaptionskip}{0pt}
    \caption{{Results on multi-image and video deep-research benchmarks.} 
    }
    \label{tab:multimodal-bench-main}
    \centering
    \scriptsize
    \setlength{\tabcolsep}{2.8pt}
    \renewcommand{\arraystretch}{1.08}
    \resizebox{0.98\textwidth}{!}{%
    \begin{tabular}{l c ccccccc ccccccc}
    \toprule
    & \textbf{VideoDR}
    & \multicolumn{7}{c}{\textbf{OneSearch-MI-Bench}}
    & \multicolumn{7}{c}{\textbf{OneSearch-Video-Bench}} \\
    \cmidrule(lr){2-2}
    \cmidrule(lr){3-9}
    \cmidrule(lr){10-16}
    \textbf{Model}
    & \textbf{All}
    & \textbf{SA}
    & \textbf{MH}
    & \textbf{Count}
    & \textbf{Join}
    & \textbf{Arith.}
    & \textbf{Comp.}
    & \textbf{All}
    & \textbf{SA}
    & \textbf{MH}
    & \textbf{Count}
    & \textbf{Join}
    & \textbf{Arith.}
    & \textbf{Comp.}
    & \textbf{All} \\
    \midrule
    \rowcolor{blue!5}
    \multicolumn{16}{c}{\textbf{Direct Reasoning}} \\

    GPT-4o~\citep{openai2024gpt4ocard}
    & \underline{42.0} & \underline{35.7} & 31.3 & 56.1 & \textbf{46.6} & \underline{23.0} & \textbf{44.9} & 39.9
    & \underline{36.0} & 25.5 & 29.5 & \textbf{30.8} & 13.3 & 29.9 & 27.4 \\

    Gemini-2.5-Flash~\citep{comanici2025gemini}
    & 39.0 & 32.1 & \underline{33.3} & \underline{61.4} & \textbf{46.6} & 21.3 & \underline{42.9} & \underline{40.2}
    & \textbf{44.0} & \underline{27.7} & \underline{32.8} & \underline{23.1} & \textbf{20.0} & \underline{33.8} & \underline{29.6} \\

    Gemini-2.5-Pro~\citep{comanici2025gemini}
    & \textbf{50.0} & \textbf{42.9} & \textbf{39.6} & \underline{61.4} & \textbf{46.6} & \textbf{34.4} & \textbf{44.9} & \textbf{45.2}
    & \textbf{44.0} & \textbf{34.0} & \textbf{37.7} & \underline{23.1} & \underline{17.8} & \textbf{37.7} & \textbf{32.3} \\

    Qwen2.5-VL-3B~\citep{bai2025qwen3vl}
    & 9.0 & 10.7 & 12.5 & 35.1 & 13.8 & 1.6 & 14.3 & 15.0
    & 12.0 & 6.4 & 18.0 & 5.8 & 4.4 & 3.9 & 8.1 \\

    Qwen2.5-VL-7B~\citep{bai2025qwen3vl}
    & 5.0 & 17.9 & 12.5 & 22.8 & 17.2 & 4.9 & 14.3 & 14.6
    & 4.0 & 4.3 & 16.4 & 5.8 & 2.2 & 6.6$^{*}$ & 7.2$^{*}$ \\

    Qwen2.5-VL-72B~\citep{bai2025qwen3vl}
    & 16.0 & 14.3 & 14.6 & 43.9 & 24.1 & 6.6 & 22.5 & 21.6
    & 28.0 & 6.4 & \underline{32.8} & 13.5 & 2.2 & 13.0 & 15.6 \\

    Qwen3-VL-8B~\citep{bai2025qwen3vl}
    & 11.0 & 14.3 & 10.4 & 26.3 & 22.4 & 3.3 & 18.4 & 16.0
    & 12.0 & 2.1 & 23.0 & 11.5 & 2.2 & 13.0 & 11.4 \\

    Qwen3-VL-30B-A3B~\citep{bai2025qwen3vl}
    & 14.0 & 21.4 & 18.8 & \textbf{68.4} & 25.9 & 11.5 & 26.5 & 29.6
    & 16.0 & 8.5 & 29.5 & 13.5 & 4.4 & 14.3 & 15.0 \\

    Qwen3-VL-32B~\citep{bai2025qwen3vl}
    & 17.0 & 25.0 & 27.1 & 42.1 & \underline{37.9} & 18.0 & 30.6 & 30.6
    & 20.0 & 6.4 & \textbf{37.7} & 9.6 & 2.2 & 20.8 & 17.3 \\

    \midrule
    \rowcolor{blue!5}
    \multicolumn{16}{c}{\textbf{Agentic Workflow}} \\

    Qwen3-VL-8B~\citep{bai2025qwen3vl}
    & 30.0 & 39.3 & 27.1 & 45.6 & 44.8 & 18.0 & \underline{40.8} & 35.6
    & \underline{40.0} & 12.8 & 22.9 & \underline{15.4} & 2.2 & 20.8 & 17.9 \\

    OpenSearch-VL-8B~\citep{chen2026opensearchvl}
& \underline{47.0}
& \textbf{57.1} & \underline{35.4} & \underline{63.2} & \underline{55.2} & \underline{24.6} & 36.7 & \underline{44.5}
& \textbf{44.0} & \underline{34.0} & \underline{23.0} & \underline{15.4} & \underline{11.1} & \underline{28.6} & \underline{24.7} \\

    \rowcolor{purple!12}
    \textbf{OneSearch-VL-8B (Ours)}
    & \textbf{57.0} & \underline{46.4} & \textbf{43.8} & \textbf{75.4} & \textbf{56.9} & \textbf{44.3} & \textbf{63.3} & \textbf{55.8}
    & \textbf{44.0} & \textbf{38.3} & \textbf{36.1} & \textbf{34.6} & \textbf{22.2} & \textbf{39.0} & \textbf{35.5} \\

    \bottomrule
    \end{tabular}%
    }
    \vspace{-2em}
\end{table}

\subsection{Ablation Studies}

\paragraph{\normalfont\textbf{SFT data mixture.}}
Table~\ref{tab:ablation}(a) compares SFT mixtures composed of single-image (I), multi-image (M), and video (V) trajectories using the same number of training iterations. Training on any one trajectory type raises the six-benchmark average from 40.7 to 51.7--55.1. The video-only model attains the strongest single-type average of 55.1 while also improving all three single-image benchmarks, indicating that research trajectories collected for one visual input type can provide useful supervision for others.

Combining trajectory types further improves overall performance. Adding either multi-image or video trajectories to the single-image data raises the average score. Joint training on all three types reaches 55.8 and gives the best result in this ablation on SimpleVQA, OneSearch-MI-Bench, and OneSearch-Video-Bench. This result supports the complementarity of the three trajectory sources under our training setup.

\paragraph{\normalfont\textbf{RL reward.}}
Table~\ref{tab:ablation}(b) analyzes answer correctness ($R_{\mathrm{acc}}$), query quality ($R_{\mathrm{query}}$), and the Evidence Traceability ($r_{\mathrm{trace}}$) and Visual Grounding ($r_{\mathrm{ground}}$) dimensions of EVGR. Starting from the joint SFT model, accuracy-only RL improves the six-benchmark average from 55.8 to 56.3, and adding query quality raises it to 57.3. Adding Trace or Ground separately further improves the average to 59.0 and 59.3, showing that each process dimension supplies useful supervision beyond answer and query rewards.

Using both Trace and Ground yields the best average of 61.1, improving by 3.8 points over answer-plus-query rewards and by 2.1 and 1.8 points over the two single-dimension variants. The full reward performs best on five benchmarks and ties for best on the remaining one. This result indicates that evidence traceability and visual grounding provide complementary feedback on factual support and the use of visual cues during research.

\begin{table}[t]
    \setlength{\abovecaptionskip}{2pt}
    \setlength{\belowcaptionskip}{2pt}
    \caption{Ablation studies on the SFT data mixture and RL reward.}
    \label{tab:ablation}
    \centering
    \footnotesize
    \setlength{\tabcolsep}{3.2pt}
    \renewcommand{\arraystretch}{1.12}
    \resizebox{\linewidth}{!}{%
    \begin{tabular}{l cccc ccc ccc c}
    \toprule
    \multicolumn{12}{c}{\textbf{(a) SFT data mixture}} \\
    \midrule
    \textbf{Data Mixture}
    & \textbf{Image}
    & \textbf{Multi-Image}
    & \multicolumn{2}{c}{\textbf{Video}}
    & \textbf{SimpleVQA}
    & \textbf{InfoSeek}
    & \textbf{FVQA}
    & \textbf{\shortstack{OneSearch-\\MI}}
    & \textbf{\shortstack{OneSearch-\\Video}}
    & \textbf{VideoDR}
    & \textbf{Avg.} \\
    \midrule
    Qwen3-VL-8B
    & $\times$ & $\times$ & \multicolumn{2}{c}{$\times$}
    & 52.0 & 50.3 & 58.7 & 35.5 & 17.9 & 30.0 & 40.7 \\
    \midrule
    I
    & $\checkmark$ & $\times$ & \multicolumn{2}{c}{$\times$}
    & 66.1 & 62.4 & 65.3 & 44.5 & 24.8 & 47.0 & 51.7 \\
    M
    & $\times$ & $\checkmark$ & \multicolumn{2}{c}{$\times$}
    & 67.0 & 62.0 & 67.1 & 48.5 & 25.0 & 49.0 & 53.1 \\
    V
    & $\times$ & $\times$ & \multicolumn{2}{c}{$\checkmark$}
    & 68.6 & 62.5 & 68.5 & 51.2 & 28.0 & 52.0 & 55.1 \\
    I + M
    & $\checkmark$ & $\checkmark$ & \multicolumn{2}{c}{$\times$}
    & 68.4 & 64.5 & 67.1 & 50.8 & 28.6 & 50.0 & 54.9 \\
    I + V
    & $\checkmark$ & $\times$ & \multicolumn{2}{c}{$\checkmark$}
    & 68.1 & 63.8 & 68.2 & 49.1 & 29.3 & 51.0 & 54.9 \\
    I + M + V
    & $\checkmark$ & $\checkmark$ & \multicolumn{2}{c}{$\checkmark$}
    & 68.7 & 64.4 & 67.8 & 51.5 & 31.2 & 51.0 & 55.8 \\
    \midrule
    \multicolumn{12}{c}{\textbf{(b) RL reward}} \\
    \midrule
    \textbf{Method}
    & $\bm{R}_{\mathbf{acc}}$
    & $\bm{R}_{\mathbf{query}}$
    & $\bm{r}_{\mathbf{trace}}$
    & $\bm{r}_{\mathbf{ground}}$
    & \textbf{SimpleVQA}
    & \textbf{InfoSeek}
    & \textbf{FVQA}
    & \textbf{\shortstack{OneSearch-\\MI}}
    & \textbf{\shortstack{OneSearch-\\Video}}
    & \textbf{VideoDR}
    & \textbf{Avg.} \\
    \midrule
    OneSearch-VL-8B-SFT
    & -- & -- & -- & --
    & 68.7 & 64.4 & 67.8 & 51.5 & 31.2 & 51.0 & 55.8 \\
    \midrule
    RL (Acc.)
    & $\checkmark$ & $\times$ & $\times$ & $\times$
    & 69.2 & 64.5 & 69.3 & 50.8 & 32.8 & 51.0 & 56.3 \\
    RL (Acc. + Query)
    & $\checkmark$ & $\checkmark$ & $\times$ & $\times$
    & 70.0 & 65.8 & 70.2 & 52.5 & 33.2 & 52.0 & 57.3 \\
    RL (Acc. + Query + Trace)
    & $\checkmark$ & $\checkmark$ & $\checkmark$ & $\times$
    & 71.0 & 68.0 & 71.5 & 54.0 & 35.5 & 54.0 & 59.0 \\
    RL (Acc. + Query + Ground)
    & $\checkmark$ & $\checkmark$ & $\times$ & $\checkmark$
    & 71.2 & 68.2 & 71.8 & 54.2 & 35.2 & 55.0 & 59.3 \\
    RL (Full reward)
    & $\checkmark$ & $\checkmark$ & $\checkmark$ & $\checkmark$
    & 72.4 & 72.3 & 73.4 & 55.8 & 35.5 & 57.0 & 61.1 \\
    \bottomrule
    \end{tabular}%
    }
\end{table}

\section{Related Work}
\label{app:related-work}

\subsection{Multimodal Search and Deep Research}

Active search enables language models to supplement parametric knowledge with external evidence acquired through multi-turn interaction \citep{jin2025searchr1trainingllmsreason}. Multimodal research extends this paradigm to visual inputs by combining image and text retrieval with visual inspection, and training search policies through synthetic tool-use trajectories and reinforcement learning \citep{wu2025mmsearchr1,geng2025webwatcher,zhou2026visiondeepresearch,chen2026opensearchvl,yao2026mmdeepresearch,narayan2025deepmmsearchr1}. Related work explores active visual perception, generalizable tool use, and joint optimization of reasoning and search, advancing research agents that integrate visual understanding with external evidence acquisition \citep{liu2025visualarft,hong2025deepeyesv2,chng2025sensenova}.

\subsection{Video Deep Research}

Video deep research additionally requires agents to locate relevant moments, identify objects across frames, and connect them to external knowledge. VideoDR formalizes this setting through an open-web benchmark \citep{liu2026videodr}. Recent methods develop video-oriented research pipelines by jointly training temporal localization, spatial inspection, and multimodal retrieval \citep{gao2026videosearcher}, or by staging tool use to guide visual grounding before web exploration \citep{fang2026videodeepresearch}. These studies highlight temporal and spatial grounding as essential links between video content and external evidence.

\subsection{Unified Visual Reasoning}

Reinforcement learning has improved image question answering and detection \citep{huang2025vision,shen2025vlm}, image segmentation \citep{you2025seg}, video question answering \citep{feng2025video}, and temporal or spatio-temporal video understanding \citep{wang2025time,li2025videochat}, with many methods designed for specific tasks. LLaVA-ST jointly addresses fine-grained spatial, temporal, and spatio-temporal understanding \citep{li2025llavast}. OneThinker jointly learns question answering, captioning, grounding, tracking, and segmentation across images and videos, demonstrating the potential of training across tasks and visual inputs \citep{feng2025onethinker}. Our work focuses on unification at the level of deep research: a single policy combines input-specific visual operations with external retrieval and fact composition across single images, image collections, and videos.

\subsection{Structural References and Process Supervision}

Beyond answer correctness, research agents receive supervision through query-quality and tool-use rewards \citep{chen2026opensearchvl,gao2026videosearcher}. Other approaches use intermediate entity anchors and citation-supported factual requirements as references for credit assignment \citep{searches2026searchworld,zhang2026carr}, while process-oriented objectives evaluate search decisions or complete rollouts \citep{yan2026prommsearchagent,wang2026drmmsearchagent}. These studies establish the value of structured evidence for agent supervision. We use VGEG to preserve visual provenance and answer-composition dependencies in a shared task-level reference, connecting task construction and verification with trajectory supervision. EVGR derives evidence-traceability and visual-grounding criteria from this reference to evaluate complete research trajectories.

\section{Conclusion}
We introduced OneSearch-VL, a unified agent for deep research over single-image, multi-image, and videos. VGEG links visual anchors, source-supported facts, and answer-producing operations, enabling verified task construction, expert-trajectory filtering, and process-level reward design. Building on this structure, EVGR evaluates both evidence traceability and visual grounding. Across seven established single-image benchmarks and two operation-oriented benchmarks, OneSearch-VL consistently improves over the evaluated open-source baselines; the ablations further demonstrate complementary supervision across visual input types and the benefit of EVGR.

\section{Limitations and Future Work}
\label{sec:limitations}
OneSearch-VL depends on external search tools such as \textsc{TextSearch} and \textsc{ImageSearch} (Table~\ref{tab:tool-environment}) and changing webpages, which can affect evidence availability and exact reproducibility. Future work should preserve retrieval snapshots and evaluate robustness to tool failures. Automated VGEG construction and model-based rewards may also introduce annotation errors or judging biases. Human calibration and open multimodal process judges could improve the reliability of trajectory-level assessment. Multi-turn interaction incurs additional inference and tool costs, motivating adaptive budgets and cost-aware training.

\clearpage
\bibliographystyle{iclr2027_conference}
\bibliography{onesearchvl}

@article{wu2025mmsearchr1,
  title   = {{MMSearch-R1}: Incentivizing {LMMs} to Search},
  author  = {Wu, Jinming and Deng, Zihao and Li, Wei and Liu, Yiding and You, Bo and Li, Bo and Ma, Zejun and Liu, Ziwei},
  journal = {arXiv preprint arXiv:2506.20670},
  year    = {2025}
}

@article{geng2025webwatcher,
  title   = {{WebWatcher}: Breaking New Frontier of Vision-Language Deep Research Agent},
  author  = {Geng, Xinyu and Xia, Peng and Zhang, Zhen and Wang, Xinyu and Wang, Qiuchen and Ding, Ruixue and Wang, Chenxi and Wu, Jialong and Zhao, Yida and Li, Kuan and Jiang, Yong and Xie, Pengjun and Huang, Fei and Zhou, Jingren},
  journal = {arXiv preprint arXiv:2508.05748},
  year    = {2025}
}

@article{zhou2026visiondeepresearch,
  title   = {{Vision-DeepResearch}: Incentivizing DeepResearch Capability in Multimodal Large Language Models},
  author  = {Huang, Wenxuan and Zeng, Yu and Wang, Qiuchen and Fang, Zhen and Cao, Shaosheng and Chu, Zheng and Yin, Qingyu and Chen, Shuang and Yin, Zhenfei and Chen, Lin and Chen, Zehui and Tang, Xu and Hu, Yao and Lin, Shaohui and Torr, Philip and Zhao, Feng and Ouyang, Wanli},
  journal = {arXiv preprint arXiv:2601.22060},
  year    = {2026}
}

@article{chen2026opensearchvl,
  title   = {{OpenSearch-VL}: An Open Recipe for Frontier Multimodal Search Agents},
  author  = {Chen, Shuang and Feng, Kaituo and Chen, Hangting and Huang, Wenxuan and Dai, Dasen and Shou, Quanxin and Lin, Yunlong and Yue, Xiangyu and Gao, Shenghua and Pang, Tianyu},
  journal = {arXiv preprint arXiv:2605.05185},
  year    = {2026}
}

@article{yao2026mmdeepresearch,
  title   = {{MM-DeepResearch}: A Simple and Effective Multimodal Agentic Search Baseline},
  author  = {Yao, Huanjin and Yin, Qixiang and Yang, Min and Zhao, Ziwang and Wang, Yibo and Luo, Haotian and Zhang, Jingyi and Huang, Jiaxing},
  journal = {arXiv preprint arXiv:2603.01050},
  year    = {2026}
}

@article{searches2026searchworld,
  title   = {{SearchEyes}: Towards Frontier Multimodal Deep Search Intelligence via Search World Simulation},
  author  = {Jiao, Zhengbo and Cheng, Yiming and Jiang, Yilei and Feng, Kaituo and Huang, Rui and Jiang, Tianyi and Tian, Juanxi and Li, Jiapeng and Wang, Qunzhong and Chen, Tailai and Wei, Qianshan and Xiao, Chuan and Rong, Shanyu and Li, Yangfu and Zhou, Yanhan and Ma, Yunpu and Zhang, Yifan and Yue, Xiangyu},
  journal = {arXiv preprint arXiv:2607.05943},
  year    = {2026}
}

@article{zhang2026carr,
  title   = {Chaining the Evidence: Robust Reinforcement Learning for Deep Search Agents with Citation-Aware Rubric Rewards},
  author  = {Zhang, Jiajie and Lv, Xin and Feng, Ling and Hou, Lei and Li, Juanzi},
  journal = {arXiv preprint arXiv:2601.06021},
  year    = {2026}
}

@article{liu2026videodr,
  title   = {Watching, Reasoning, and Searching: A Video Deep Research Benchmark on Open Web for Agentic Video Reasoning},
  author  = {Liu, Chengwen and Yu, Xiaomin and Chang, Zhuoyue and Huang, Zhe and Zhang, Shuo and Lian, Heng and Wang, Kunyi and Xu, Rui and Hu, Sen and Hou, Jianheng and Peng, Hao and Qin, Chengwei and Hu, Xiaobin and Peng, Hong and Chen, Ronghao and Wang, Huacan},
  journal = {arXiv preprint arXiv:2601.06943},
  year    = {2026}
}

@article{gao2026videosearcher,
  title   = {{VideoSearcher}: Empowering Video Deep Research with Multi-Tool Agentic Reasoning via Reinforcement Learning},
  author  = {Gao, Zhenkun and Bao, Yicheng and Peng, Jinlong and Li, Xueheng and Huang, Theo and Liu, Bangwei and Li, Kunquan and Gan, Zhenye and Hu, Tao and Xie, Chengjun and Yang, Mingqian and He, Xuanhua and Zhang, Zhizhong and Tan, Xin and Wang, Chengjie and Xie, Yuan},
  journal = {arXiv preprint arXiv:2607.02927},
  year    = {2026}
}

@article{fang2026videodeepresearch,
  title   = {{Video-DeepResearch}: Towards the Next-Generation Multimodal Deepresearch Agent},
  author  = {Fang, Zhen and Zeng, Yu and Huang, Wenxuan and Zhao, Yiming and Huang, Shiting and Ren, Tianfei and Lu, Qi and Ren, Qingnan and Su, Qisheng and Wang, Lionel Z. and Yin, Qingyu and Chen, Shuang and Chen, Zehui and Chen, Lin and Yin, Zhenfei and Hu, Yao and Lin, Shaohui and Ouyang, Wanli and Cao, Shaosheng and Zhao, Feng},
  journal = {arXiv preprint arXiv:2608.03979},
  year    = {2026}
}

@inproceedings{li2025llavast,
  title     = {{LLaVA-ST}: A Multimodal Large Language Model for Fine-Grained Spatial-Temporal Understanding},
  author    = {Li, Hongyu and Chen, Jinyu and Wei, Ziyu and Huang, Shaofei and Hui, Tianrui and Gao, Jialin and Wei, Xiaoming and Liu, Si},
  booktitle = {Proceedings of the IEEE/CVF Conference on Computer Vision and Pattern Recognition (CVPR)},
  month     = {June},
  year      = {2025},
  pages     = {8592--8603}
}

@article{feng2025onethinker,
  title   = {{OneThinker}: All-in-One Reasoning Model for Image and Video},
  author  = {Feng, Kaituo and Zhang, Manyuan and Li, Hongyu and Fan, Kaixuan and Chen, Shuang and Jiang, Yilei and Zheng, Dian and Sun, Peiwen and Zhang, Yiyuan and Sun, Haoze and Feng, Yan and Pei, Peng and Cai, Xunliang and Yue, Xiangyu},
  journal = {arXiv preprint arXiv:2512.03043},
  year    = {2025}
}

@article{yan2026prommsearchagent,
  title   = {{ProMMSearchAgent}: A Generalizable Multimodal Search Agent Trained with Process-Oriented Rewards},
  author  = {Yan, Wentao and Wang, Shengqin and Zhou, Huichi and Chen, Yihang and Shao, Kun and Xie, Yuan and Zhang, Zhizhong},
  journal = {arXiv preprint arXiv:2604.20486},
  year    = {2026}
}

@article{wang2026drmmsearchagent,
  title   = {{DR-MMSearchAgent}: Deepening Reasoning in Multimodal Search Agents},
  author  = {Wang, Shengqin and Yan, Wentao and Zhou, Huichi and Chen, Yihang and Shao, Kun and Zhang, Zhizhong and Xie, Yuan},
  journal = {arXiv preprint arXiv:2604.19264},
  year    = {2026}
}

@misc{openai2024gpt4ocard,
  title         = {{GPT-4o} System Card},
  author        = {{OpenAI Team}},
  year          = {2024},
  eprint        = {2410.21276},
  archiveprefix = {arXiv},
  primaryclass  = {cs.CL}
}

@misc{openai2025gpt5,
  author       = {{OpenAI}},
  title        = {Introducing {GPT-5}},
  howpublished = {\url{https://openai.com/index/introducing-gpt-5/}},
  year         = {2025}
}

@article{comanici2025gemini,
  title   = {{Gemini 2.5}: Pushing the Frontier with Advanced Reasoning, Multimodality, Long Context, and Next Generation Agentic Capabilities},
  author  = {Comanici, Gheorghe and Bieber, Eric and Schaekermann, Mike and Pasupat, Ice and Sachdeva, Noveen and Dhillon, Inderjit and Blistein, Marcel and Ram, Ori and Zhang, Dan and Rosen, Evan and others},
  journal = {arXiv preprint arXiv:2507.06261},
  year    = {2025}
}

@misc{anthropic2025claude4,
  author       = {{Anthropic Team}},
  title        = {System Card: {Claude Opus 4} and {Claude Sonnet 4}},
  howpublished = {\url{https://www-cdn.anthropic.com/6d8a8055020700718b0c49369f60816ba2a7c285.pdf}},
  year         = {2025}
}

@misc{anthropic2025claude37,
  author       = {{Anthropic Team}},
  title        = {{Claude 3.7 Sonnet} System Card},
  howpublished = {\url{https://www-cdn.anthropic.com/9ff93dfa8f445c932415d335c88852ef47f1201e.pdf}},
  year         = {2025}
}

@article{bai2025qwen3vl,
  title   = {{Qwen3-VL} Technical Report},
  author  = {Bai, Shuai and Cai, Yuxuan and Chen, Ruizhe and Chen, Keqin and Chen, Xionghui and Cheng, Zesen and Deng, Lianghao and Ding, Wei and Gao, Chang and Ge, Chunjiang and others},
  journal = {arXiv preprint arXiv:2511.21631},
  year    = {2025}
}

@article{narayan2025deepmmsearchr1,
  title   = {{DeepMMSearch-R1}: Empowering Multimodal {LLMs} in Multimodal Web Search},
  author  = {Narayan, Kartik and Xu, Yang and Cao, Tian and Nerella, Kavya and Patel, Vishal M. and Shiee, Navid and Grasch, Peter and Jia, Chao and Yang, Yinfei and Gan, Zhe},
  journal = {arXiv preprint arXiv:2510.12801},
  year    = {2025}
}

@article{liu2025visualarft,
  title   = {Visual Agentic Reinforcement Fine-Tuning},
  author  = {Liu, Ziyu and Zang, Yuhang and Zou, Yushan and Liang, Zijian and Dong, Xiaoyi and Cao, Yuhang and Duan, Haodong and Lin, Dahua and Wang, Jiaqi},
  journal = {arXiv preprint arXiv:2505.14246},
  year    = {2025}
}

@article{hong2025deepeyesv2,
  title   = {{DeepEyesV2}: Toward Agentic Multimodal Model},
  author  = {Hong, Jack and Zhao, Chenxiao and Zhu, ChengLin and Lu, Weiheng and Xu, Guohai and Yu, Xing},
  journal = {arXiv preprint arXiv:2511.05271},
  year    = {2025}
}

@article{chng2025sensenova,
  title   = {{SenseNova-MARS}: Empowering Multimodal Agentic Reasoning and Search via Reinforcement Learning},
  author  = {Chng, Yong Xien and Hu, Tao and Tong, Wenwen and Li, Xueheng and Chen, Jiandong and Yu, Haojia and Lu, Jiefan and Guo, Hewei and Deng, Hanming and Xie, Chengjun and others},
  journal = {arXiv preprint arXiv:2512.24330},
  year    = {2025}
}

@article{xu2026stamp,
  title   = {{STAMP}: Provenance-Guided Credit Assignment for Deep Search Agents},
  author  = {Xu, Ke and Xu, Han and Chen, Xinran and Wang, Yuqian and Li, Zhixuan and Liu, Xiaojian and Wu, Changwo and Xia, Jianqiang and Li, Yuchen},
  journal = {arXiv preprint arXiv:2607.11172},
  year    = {2026}
}

@article{sun2026hievidr,
  title   = {{HiEviDR-Bench}: A Benchmark for Hierarchical Evidence Aggregation in Deep Research},
  author  = {Sun, Yubo and Peng, Chunyi and Yan, Yukun and Liu, Zhenghao and Mei, Sen and Xu, Bangrui and Zhou, Xuanhe and Chen, Chi and Sun, Maosong},
  journal = {arXiv preprint arXiv:2607.25151},
  year    = {2026}
}

@inproceedings{cheng2025simplevqa,
  title     = {{SimpleVQA}: Multimodal Factuality Evaluation for Multimodal Large Language Models},
  author    = {Cheng, Xianfu and Zhang, Wei and Zhang, Shiwei and Yang, Jian and Guan, Xiangyuan and Wu, Xianjie and Li, Xiang and Zhang, Ge and Liu, Jiaheng and Mai, Yuying and others},
  booktitle = {Proceedings of the IEEE/CVF International Conference on Computer Vision},
  pages     = {4637--4646},
  year      = {2025}
}

@article{vdr,
  title   = {{Vision-DeepResearch Benchmark}: Rethinking Visual and Textual Search for Multimodal Large Language Models},
  author  = {Zeng, Yu and Huang, Wenxuan and Fang, Zhen and Chen, Shuang and Shen, Yufan and Cai, Yishuo and Wang, Xiaoman and Yin, Zhenfei and Chen, Lin and Chen, Zehui and Huang, Shiting and Zhao, Yiming and Hu, Yao and Torr, Philip and Ouyang, Wanli and Cao, Shaosheng},
  journal = {arXiv preprint arXiv:2602.02185},
  year    = {2026}
}

@inproceedings{mmsearch,
  title     = {{MMSearch}: Benchmarking the Potential of Large Models as Multi-modal Search Engines},
  author    = {Jiang, Dongzhi and Zhang, Renrui and Guo, Ziyu and Wu, Yanmin and Lei, Jiayi and Qiu, Pengshuo and Lu, Pan and Chen, Zehui and Fu, Chaoyou and Song, Guanglu and Gao, Peng and Liu, Yu and Li, Chunyuan and Li, Hongsheng},
  booktitle = {International Conference on Learning Representations},
  year      = {2025}
}

@article{fu2025livevqa,
  title   = {{LiveVQA}: Live Visual Knowledge Seeking},
  author  = {Fu, Mingyang and Peng, Yuyang and Liu, Benlin and Wan, Yao and Chen, Dongping},
  journal = {arXiv preprint arXiv:2504.05288},
  year    = {2025}
}

@article{webwatcher,
  title   = {{WebWatcher}: Breaking New Frontiers of Vision-Language Deep Research Agent},
  author  = {Geng, Xinyu and Xia, Peng and Zhang, Zhen and Wang, Xinyu and Wang, Qiuchen and Ding, Ruixue and Wang, Chenxi and Wu, Jialong and Zhao, Yida and Li, Kuan and Jiang, Yong and Xie, Pengjun and Huang, Fei and Zhou, Jingren},
  journal = {arXiv preprint arXiv:2508.05748},
  year    = {2025}
}

@article{wang2017fvqa,
  title     = {{FVQA}: Fact-Based Visual Question Answering},
  author    = {Wang, Peng and Wu, Qi and Shen, Chunhua and Dick, Anthony and Van Den Hengel, Anton},
  journal   = {IEEE Transactions on Pattern Analysis and Machine Intelligence},
  volume    = {40},
  number    = {10},
  pages     = {2413--2427},
  year      = {2017},
  publisher = {IEEE}
}

@article{chen2023can,
  title   = {Can Pre-trained Vision and Language Models Answer Visual Information-Seeking Questions?},
  author  = {Chen, Yang and Hu, Hexiang and Luan, Yi and Sun, Haitian and Changpinyo, Soravit and Ritter, Alan and Chang, Ming-Wei},
  journal = {arXiv preprint arXiv:2302.11713},
  year    = {2023}
}

@article{jin2025searchr1trainingllmsreason,
  title   = {{Search-R1}: Training {LLMs} to Reason and Leverage Search Engines with Reinforcement Learning},
  author  = {Jin, Bowen and Zeng, Hansi and Yue, Zhenrui and Yoon, Jinsung and Arik, Sercan and Wang, Dong and Zamani, Hamed and Han, Jiawei},
  journal = {arXiv preprint arXiv:2503.09516},
  year    = {2025}
}

@article{huang2025vision,
  title   = {{Vision-R1}: Incentivizing Reasoning Capability in Multimodal Large Language Models},
  author  = {Huang, Wenxuan and Jia, Bohan and Zhai, Zijie and Cao, Shaosheng and Ye, Zheyu and Zhao, Fei and Xu, Zhe and Hu, Yao and Lin, Shaohui},
  journal = {arXiv preprint arXiv:2503.06749},
  year    = {2025}
}

@article{shen2025vlm,
  title   = {{VLM-R1}: A Stable and Generalizable R1-Style Large Vision-Language Model},
  author  = {Shen, Haozhan and Liu, Peng and Li, Jingcheng and Fang, Chunxin and Ma, Yibo and Liao, Jiajia and Shen, Qiaoli and Zhang, Zilun and Zhao, Kangjia and Zhang, Qianqian and others},
  journal = {arXiv preprint arXiv:2504.07615},
  year    = {2025}
}

@article{you2025seg,
  title   = {{Seg-R1}: Segmentation Can Be Surprisingly Simple with Reinforcement Learning},
  author  = {You, Zuyao and Wu, Zuxuan},
  journal = {arXiv preprint arXiv:2506.22624},
  year    = {2025}
}

@article{feng2025video,
  title   = {{Video-R1}: Reinforcing Video Reasoning in Multimodal Large Language Models},
  author  = {Feng, Kaituo and Gong, Kaixiong and Li, Bohao and Guo, Zonghao and Wang, Yibing and Peng, Tianshuo and Wu, Junfei and Zhang, Xiaoying and Wang, Benyou and Yue, Xiangyu},
  journal = {arXiv preprint arXiv:2503.21776},
  year    = {2025}
}

@article{wang2025time,
  title   = {{Time-R1}: Post-Training Large Vision Language Model for Temporal Video Grounding},
  author  = {Wang, Ye and Wang, Ziheng and Xu, Boshen and Du, Yang and Lin, Kejun and Xiao, Zihan and Yue, Zihao and Ju, Jianzhong and Zhang, Liang and Yang, Dingyi and others},
  journal = {arXiv preprint arXiv:2503.13377},
  year    = {2025}
}

@article{li2025videochat,
  title   = {{VideoChat-R1}: Enhancing Spatio-Temporal Perception via Reinforcement Fine-Tuning},
  author  = {Li, Xinhao and Yan, Ziang and Meng, Desen and Dong, Lu and Zeng, Xiangyu and He, Yinan and Wang, Yali and Qiao, Yu and Wang, Yi and Wang, Limin},
  journal = {arXiv preprint arXiv:2504.06958},
  year    = {2025}
}

@article{li2026qwen3vlembedding,
  title   = {{Qwen3-VL-Embedding} and {Qwen3-VL-Reranker}: A Unified Framework for State-of-the-Art Multimodal Retrieval and Ranking},
  author  = {Li, Mingxin and Zhang, Yanzhao and Long, Dingkun and Chen, Keqin and Song, Sibo and Bai, Shuai and Yang, Zhibo and Xie, Pengjun and Yang, An and Liu, Dayiheng and Zhou, Jingren and Lin, Junyang},
  journal = {arXiv preprint arXiv:2601.04720},
  year    = {2026}
}

@misc{openai2025gpt41,
  author       = {{OpenAI}},
  title        = {Introducing {GPT-4.1} in the {API}},
  howpublished = {\url{https://openai.com/index/gpt-4-1/}},
  year         = {2025}
}

@article{bytedanceseed2026seed20,
  title   = {{Seed2.0} Model Card: Towards Intelligence Frontier for Real-World Complexity},
  author  = {{ByteDance Seed}},
  journal = {arXiv preprint arXiv:2607.00248},
  year    = {2026}
}

@article{yang2025qwen3,
  title   = {{Qwen3} Technical Report},
  author  = {Yang, An and Li, Anfeng and Yang, Baosong and Zhang, Beichen and Hui, Binyuan and Zheng, Bo and Yu, Bowen and Gao, Chang and Huang, Chengen and Lv, Chenxu and others},
  journal = {arXiv preprint arXiv:2505.09388},
  year    = {2025}
}

\clearpage
\appendix
\titleformat*{\paragraph}{\normalfont\sffamily\bfseries}
\etocdepthtag.toc{appendix}
\begingroup
  \etocsettagdepth{main}{none}
  \etocsettagdepth{appendix}{subsection}
  \etocsetnexttocdepth{subsection}
  \etocsettocstyle{\section*{Appendix Contents}}{}
  \tableofcontents
\endgroup
\clearpage
\section{Details of the VGEG-Centered Multimodal Data Engine}
\label{app:data-engine}

This appendix expands the five stages in Figure~\ref{fig:data-pipeline}.
The input-level graph $\mathcal G_X$ collects candidate visual and web
evidence for a visual source. A task-level VGEG $\Gamma_i$ is derived
from $\mathcal G_X$ through a task-conditioned projection: it selects
the anchors and facts needed by a particular question and augments them
with explicit answer-producing operations and dependencies.
We report the recorded sampling and execution settings for each stage;
prompt templates are omitted.

\subsection{Stage 1: Visual Source Curation}
\label{app:source-filtering}

Figure~\ref{fig:data-pipeline}(a) contains six blocks: Web Video Source Pool, Upload Date Filter, Metadata Coarse Filter, Category Balance \& Exclusion, Video Snippet LLM Filter, and Video Duration Balance. Table~\ref{tab:source-curation} gives the recorded size after each block.

\begin{table}[ht]
\centering
\small
\caption{Source-video curation. Counts refer to candidate video records.}
\label{tab:source-curation}
\begin{tabular}{lr}
\toprule
Stage & Retained records \\
\midrule
Web video source pool & 2,500,000 \\
Upload date filter & 1,295,220 \\
Metadata coarse filter & 377,955 \\
Category balance \& exclusion & 127,562 \\
Video snippet LLM filter & 72,403 \\
Video duration balance & 70,781 \\
\bottomrule
\end{tabular}
\end{table}

\paragraph{Step 1: Web Video Source Pool.} The pipeline starts from 2.5M candidate YouTube video records. This pool provides the common input to all subsequent source-level filters.

\paragraph{Step 2: Upload Date Filter.} We retain records with available metadata and an upload date on or after January 1, 2023. This stage reduces the pool to 1.3M records.

\paragraph{Step 3: Metadata Coarse Filter.} A rule-based score combines entity and event cues in the title, description, and tags with duration and metadata richness. We retain the higher-scoring records, leaving 378k candidates.

\paragraph{Step 4: Category Balance \& Exclusion.} We retain Film \& Animation, Autos \& Vehicles, Pets \& Animals, Sports, Travel \& Events, Gaming, People \& Blogs, Comedy, Entertainment, News \& Politics, Howto \& Style, and Science \& Technology. Candidates are ranked within each category, with equal category quotas used to maintain content coverage and per-channel limits used to prevent concentration on a small number of sources. We exclude Music, Nonprofits \& Activism, Education, and Unknown because these categories more often exhibit limited visual dynamics or contain fewer identifiable, searchable entities for evidence-grounded task construction. This stage retains 128k records.

\paragraph{Step 5: Video Snippet LLM Filter.} We use GPT-4.1~\citep{openai2025gpt41} to assess whether each video potentially contains multiple scenes, searchable entities, specific events, visual information, and available web evidence. We combine its assessment with the rule-based metadata score and retain 72k candidates. This step operates on textual metadata; visual inspection occurs in the subsequent anchor-discovery stage.

\paragraph{Step 6: Video Duration Balance.} We group the remaining candidates by video duration and sample across duration ranges to reduce the dominance of any single interval. This step yields 70k records with a more balanced duration distribution.

\subsection{Stage 2: Dense Visual Anchor Discovery}
\label{app:visual-anchors}

Figure~\ref{fig:data-pipeline}(b) contains five blocks: Video Clip \& Caption, Key-Frame Selection \& Caption, Cross-Clip Frame Deduplication, Event Aggregation \& Caption, and Event Key-Frame Selection \& Key Object Grounding. Their outputs form the Video Dense Caption Tree.

\paragraph{Step 1: Video Clip \& Caption.} We uniformly sample each video at a target rate of 2 FPS, 
and divide the frames into temporally ordered local clips of ten frames. A final group with fewer than five frames is merged into its predecessor. We use Seed 2.0 Pro~\citep{bytedanceseed2026seed20} to describe the scenes, actions, and identifiable entities in each clip.

\paragraph{Step 2: Key-Frame Selection \& Caption.} For each clip, we use Seed 2.0 Pro~\citep{bytedanceseed2026seed20} to select representative key frames and generate key-moment descriptions and candidate entity cues, while retaining the frame indices and timestamps.

\paragraph{Step 3: Cross-Clip Frame Deduplication.} We use Qwen3-VL-Embedding-8B~\citep{li2026qwen3vlembedding} to encode each candidate frame and its description as normalized image and text embeddings. Candidates are compared in temporal order against the most recently retained frame. A candidate is removed when image similarity is at least 0.9 or text similarity is at least 0.8. The similarity thresholds are stored with the deduplication output.

\paragraph{Step 4: Event Aggregation \& Caption.} We use Seed 2.0 Pro~\citep{bytedanceseed2026seed20} to merge temporally adjacent, semantically related clips into coherent events based on the clip descriptions and deduplicated key frames. Each event records a title, temporal range, description, covered clip range, and searchable entity cues, forming the temporal hierarchy from the video to its events.

\paragraph{Step 5: Event Key-Frame Selection \& Key Object Grounding.} For each event, we select representative key frames and use Qwen3-VL-30B-A3B-Instruct~\citep{bai2025qwen3vl} to localize visible, searchable objects. Each object records a name or description, a type, a normalized bounding box, and its relation to the event. Bounding boxes use coordinates in $[0,1000]^4$.

\paragraph{Video Dense Caption Tree.} The resulting hierarchy preserves temporal and spatial references:
\begin{equation}
\text{video}\rightarrow\text{event}\rightarrow
\text{key frame}\rightarrow\text{localized object}.
\end{equation}
These records supply the visual anchors used by later retrieval and question construction.

\subsection{Stage 3: Web Evidence Graph Construction}
\label{app:web-evidence}

Figure~\ref{fig:data-pipeline}(c) routes localized Objects through image search or OCR followed by text search, assembles a Candidate Entity Pool, expands related entities and attributes through the Expand Entity Pipeline, and produces a Web Entity Graph.

\paragraph{Objects and Retrieval Routing.} Object grouping and representative-view selection use deterministic matching. Objects with matching types and similar entity or name cues are grouped across frames; the largest available bounding box is selected as the representative view, with alternate views retained from other frames. Crops with a short side below 256 pixels are enlarged to 256 pixels. We use GPT-4.1~\citep{openai2025gpt41} to route each object to reverse image search, OCR followed by text search, or omission based on its crop, name, type, candidate entity cues, and event relation. For the OCR route, we also use the model to read visible text and propose a candidate entity; the image-search route uses an external reverse-image-search service to retrieve candidate entities and webpages.

\paragraph{Candidate Entity Pool.} Because reverse image search is sensitive to viewpoint and cropping, we construct a retrieval pool for each object grouped across frames, containing its representative crop and alternate views from other sampled frames. If one crop yields no usable candidates, we retry the search with another view from the same object pool. We first use Seed 2.0 Pro~\citep{bytedanceseed2026seed20} to select visually grounded anchors with research potential from the event context and initial retrieval results. We then use the model to combine the object crop, local visual description, and image-search or OCR evidence to propose an entity identity, assess whether the retrieved entity matches the visual object, and formulate entity-specific text queries. We use Qwen3-32B~\citep{yang2025qwen3} to read the retrieved pages, produce query-focused summaries, and verify consistency between the page evidence and the candidate identity; search snippets are used when the page body is unavailable. Duplicate pages and explicitly mismatched identities are removed. The remaining identities, source objects, and webpage evidence form the candidate entity pool.

\paragraph{Expand Entity Pipeline.} We use Seed 2.0 Pro~\citep{bytedanceseed2026seed20} to extract source-supported relations from the verified page summaries. Each fact records its head entity, relation, tail entity or attribute value, supporting URL, and evidence quotation:
\begin{equation}
f=(u,r,z,p,\xi).
\end{equation}
Here, $p$ is the source URL and $\xi$ is the retained supporting text. A deterministic parser checks required fields and matches normalized quotations against the supplied source material; a matched quotation must contain at least six normalized characters. Numeric facts can additionally retain units and value types. For entity-valued tails, we use Seed 2.0 Pro~\citep{bytedanceseed2026seed20} to decide whether to continue based on entity specificity, relation confidence, task relevance, and evidence-expansion potential and to propose a follow-up query; we then use Qwen3-32B~\citep{yang2025qwen3} to summarize and verify the retrieved expansion pages. For each expandable entity, the pipeline retains one query, retrieves five results, deduplicates pages by host, and reads the selected pages. Each parent selects high-confidence successors with scores of at least 0.5, and expansion stops when no suitable successor remains.

\paragraph{Web Entity Graph.} Graph assembly uses no additional model and deterministically merges the preceding outputs. The graph contains events, frames, localized objects, entities, attributes, and source pages. Structural edges associate events with frames and frames with objects; grounding edges bind objects to candidate entities; factual edges connect entities to other entities or attributes while retaining source evidence. Entity and relation normalization merges compatible records, removes duplicate edges, and removes self-loops. The resulting input-level graph $\mathcal G_X$ supplies a common evidence context for multiple candidate questions.

\subsection{Stage 4: VGEG-Based Task Construction}
\label{app:vgeg-construction}

Figure~\ref{fig:data-pipeline}(d) contains four blocks: LLM-Guided QA Generation, QA Verifier, Visual Entity Fuzzing Rewrite, and Quality Filter. VGEG is retained as the task-level reference throughout these blocks.

\paragraph{Step 1: LLM-Guided QA Generation.} We organize clip and event descriptions from the dense video caption tree, visual-anchor records, source-supported facts from the input-level graph $\mathcal G_X$, and candidate relation chains into a generation context. We use Seed 2.0 Pro~\citep{bytedanceseed2026seed20} to propose candidate questions spanning the six research operations. The model uses visual descriptions and anchor mappings rather than raw video frames and outputs a question, answer, supporting facts, operation type, visual references, and corresponding frame indices. Multi-anchor questions retain a separate visual description and frame mapping for each referenced object. Relation chains can occur within each branch before the branch results are joined, counted, compared, or used in arithmetic.

The same step assembles the task-level VGEG using the main-text definition $\Gamma_i=\Phi_i(\mathcal G_X)= (\mathcal A_i,\mathcal F_i,\mathcal O_i,\mathcal R_i)$. The projection selects $\mathcal A_i$ from the visual anchors and grounded entities in $\mathcal G_X$, retaining their image, frame, or region locations. It selects $\mathcal F_i$ from the graph relations and attributes that provide source support for the task. The operation set $\mathcal O_i$ describes the answer-producing retrieval or composition, while $\mathcal R_i$ retains the selected grounding and support edges and adds task-specific dependencies connecting anchors, entities, facts, and operations. These components are assembled deterministically from the visual-reference fields, selected \texttt{used\_facts}, task-structure labels, source records, and their grounding and dependency links, without an additional model call.

\paragraph{Step 2: QA Verifier.} The verifier first checks the evidence path and task structure against the candidate graph using deterministic rules without an additional model call. Relation-chain questions are checked for continuity between successive facts. Multi-anchor questions require distinct referenced objects and corresponding supporting facts. Comparisons use a common semantic attribute; arithmetic uses numeric operands with compatible meanings; knowledge-conditioned counting associates counted entities with facts supporting the condition. Frame references are checked against the available entity-to-frame mappings, and the facts required to derive the answer must be supported by the candidate graph.

For a candidate $(q_i,y_i,\Gamma_i)$, the QA Verifier then gathers graph relations, source quotations, page summaries, and visual descriptions associated with its supporting entities. We use Seed 2.0 Pro~\citep{bytedanceseed2026seed20} to re-answer the question using only this evidence and to check semantic agreement between the regenerated and proposed answers, allowing aliases and equivalent units or date formats. To prevent answer leakage, explicit occurrences of the candidate answer in the fact summary are masked when possible. This stage checks that the supplied evidence supports recovery of the answer.

\paragraph{Step 3: Visual Entity Fuzzing Rewrite.} We use Seed 2.0 Pro~\citep{bytedanceseed2026seed20} to inspect the referenced key frames and rewrite explicit entity names into descriptions that locate the intended objects. Multi-hop questions also conceal intermediate entity names so that the final question starts from visible targets without exposing entities along the retrieval path.

The same step instantiates the task for its target visual input type. For a multi-image task, we extract the deduplicated key frames referenced by its VGEG, organize them as an unordered image collection, and remap frame- and region-level references to image indices. References to video playback, timestamps, and unnecessary temporal order are removed so that the question is self-contained over the image collection. For a video task, we retain the original video and its event, timestamp, key-frame, and region bindings, preserving the need for temporal localization.

\paragraph{Step 4: Quality Filter.} We use Seed 2.0 Pro~\citep{bytedanceseed2026seed20} to assess information masking, referential uniqueness, visual relevance, and whether external retrieval is required. A question-only probe rejects a candidate when the model can reliably recover the reference answer without the visual input. The retained record contains the visual input, final question, answer, VGEG annotations, operation label, and visual-reference mappings.

\subsection{Stage 5: Expert Trajectory Synthesis}
\label{app:trajectory-synthesis}

Figure~\ref{fig:data-pipeline}(e) contains two steps: multi-turn trajectory synthesis and rejection sampling.

\paragraph{Multi-turn trajectory synthesis.} We use Seed 2.0 Pro~\citep{bytedanceseed2026seed20} as the expert model to solve each verified task in the real tool environment. The expert receives the visual input, question, and tool definitions, while the reference answer is withheld for subsequent evaluation. At each turn, the model produces reasoning followed by a tool call or a final response. The environment executes the tool and appends its observation, including returned visual resources, to the interaction history, producing an interleaved reasoning--action--observation trajectory.

\paragraph{Rejection sampling: answer-correctness judge.} We use GPT-4o~\citep{openai2024gpt4ocard} to compare the final response of each raw expert rollout with the reference answer while allowing semantically equivalent formulations. Rollouts with an incorrect answer or without a valid terminal response are rejected.

\paragraph{Rejection sampling: process-level judge.} The remaining rollouts undergo process-level evaluation, which checks that each trajectory contains at least one effective tool call, remains logically consistent with tool observations, and avoids ineffective repetition. We retain only trajectories that pass both judges as high-quality expert demonstrations for SFT.

\section{Benchmark Construction and Evaluation}
\label{app:benchmark-details}

This appendix details the construction protocols for OneSearch-MI-Bench and OneSearch-Video-Bench, including visual-input construction, VGEG-based operation assignment, difficulty and visual-domain annotation, and human review.

\subsection{Visual Input Construction}

For OneSearch-MI-Bench, deduplicated evidence frames from the same source video form an unordered image collection. Questions are rewritten to refer to the images and visible objects, while image indices preserve the bindings to their visual anchors. References to video playback, frame numbers, and unnecessary temporal order are removed. Each retained item requires evidence from at least two images, preventing reduction to a single-image question.

For OneSearch-Video-Bench, we preserve the temporal structure of the source video and bind each visual anchor to its corresponding event and key frames. Questions refer to relevant content through visible objects, events, or temporal descriptions, requiring the model to localize visual evidence before retrieving and composing external facts.

\subsection{Research Operations and Reference Structures}
\label{app:research-operations}

Each benchmark item stores $(X_i,q_i,y_i,\Gamma_i,s_i)$, where $s_i$ is determined by the final answer-producing operation in its task-level VGEG rather than by visual content or source category. Figure~\ref{fig:benchmark-overview}(a)--(f) shows representative examples of the six operations, and Table~\ref{tab:benchmark-structure} provides their definitions and distributions.

When evidence branches contain intermediate relation chains, the category follows the operation that combines the branch results into the answer. For example, two anchors may each initiate multi-hop retrieval, but a question is labeled as multi-anchor comparison when its final answer compares the retrieved attributes. This rule assigns one principal operation to each item while preserving all intermediate dependencies in its VGEG.

The two benchmarks contain 608 questions. The five compositional categories beyond single-anchor lookup account for 555 questions, or 91.3\% of the combined set. Multi-image questions contain \mbox{2--8 images}, with a mean of 3.05 images per question. These statistics show that the benchmarks primarily test relation tracing, conditional filtering, and cross-anchor fact composition.
\begin{table}[!ht]
    \setlength{\abovecaptionskip}{2pt}
    \setlength{\belowcaptionskip}{2pt}
    \caption{Definitions and distribution of the six principal research operations.}
    \label{tab:benchmark-structure}
    \centering
    \scriptsize
    \setlength{\tabcolsep}{4.2pt}
    \renewcommand{\arraystretch}{1.10}
    \resizebox{0.85\linewidth}{!}{%
    \begin{tabular}{llrrr}
    \toprule
    \textbf{Operation} & \textbf{Definition} & \textbf{Multi-image} & \textbf{Video} & \textbf{Total} \\
    \midrule
    Single-anchor lookup & Retrieve one external attribute for one anchor. & 28 & 25 & 53 \\
    Multi-hop retrieval & Follow an external relation chain from one anchor. & 48 & 47 & 95 \\
    Knowledge-conditioned count & Test a retrieved condition per anchor, then count. & 57 & 61 & 118 \\
    Multi-anchor join & Combine retrieved facts across anchors. & 58 & 52 & 110 \\
    Multi-anchor arithmetic & Compute over retrieved numeric facts. & 61 & 45 & 106 \\
    Multi-anchor comparison & Compare the same retrieved attribute across anchors. & 49 & 77 & 126 \\
    \midrule
    \textbf{All questions} & & \textbf{301} & \textbf{307} & \textbf{608} \\
    \bottomrule
    \end{tabular}%
    }
\end{table}

\subsection{Difficulty and Visual-Domain Distributions}

We derive a structural difficulty score from the principal research operation, reasoning depth, and number of required facts. Let $h_i$ denote the number of reasoning hops, $|\mathcal F_i|$ the number of used facts in the VGEG, and $w(s_i)$ the operation weight, set to 0 for single-anchor lookup, 1 for multi-hop retrieval and knowledge-conditioned counting, and 2 for the three multi-anchor operations:
\begin{equation}
d_i=10w(s_i)+h_i+0.1|\mathcal F_i|.
\end{equation}

\begin{figure}[ht]
    \centering
    \includegraphics[width=\linewidth]{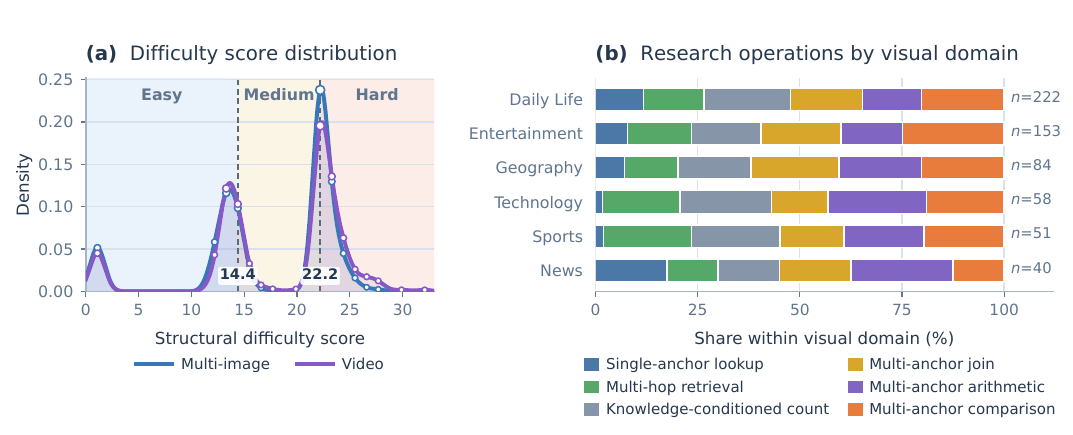}
    \caption{\textbf{Benchmark distributions.}
    (a) Smoothed structural-difficulty score distributions, with circles marking observed values. Dashed lines mark the empirical boundaries at 14.4 and 22.2; the easy, medium, and hard groups contain 100/100/101 multi-image and 102/102/103 video questions.
    (b) Operation composition within six visual domains. Bars are normalized within each domain, and $n$ denotes the number of questions. Every domain covers all six principal research operations.}
    \label{fig:benchmark-distributions}
\end{figure}

Within each benchmark, examples are ranked by this score and partitioned into easy, medium, and hard groups. Both branches yield empirical boundary scores of 14.4 and 22.2; ties at a boundary are deterministically assigned across adjacent groups to keep their sizes balanced. The three groups contain 100, 100, and 101 multi-image questions and 102, 102, and 103 video questions, respectively. Figure~\ref{fig:benchmark-distributions}(a) shows the full score distributions and their partition boundaries. These labels represent relative evidence complexity within each visual input type rather than independently calibrated human difficulty.

We divide the source videos into six broad visual domains: Daily Life, Entertainment, Geography, Technology, Sports, and News. As shown in Figure~\ref{fig:benchmark-distributions}(b), every visual domain spans all six principal research operations, indicating that the operation taxonomy is not tied to a particular source domain. Differences in their proportions reflect the benchmark's overall emphasis on compositional research questions.

\subsection{Human Review and Annotation}

Candidates first pass the automatic information-masking, referential-uniqueness, visual-relevance, and non-triviality checks described in Appendix~\ref{app:vgeg-construction}. These checks prevent answer or intermediate-entity leakage, ensure that each visual reference identifies its intended object, remove questions that do not require the visual input, and reject questions that can be answered reliably from text alone.

After automatic filtering, human annotators jointly inspect the visual input, question, reference answer, task-level VGEG, and supporting webpages. They verify that each visual reference identifies the intended object, each required fact is source-supported, the reference answer follows from the VGEG dependencies, and the principal operation matches the final answer-producing step. The reviewed record retains the confirmed visual-anchor bindings, supporting facts, reference answer, and operation annotation.

A final quality review further checks question clarity, the presence and distinguishability of the referenced visual objects, and consistency across the annotations. Candidates with ambiguous visual references, unsupported required facts, indeterminate answers, or unclear operation labels are rejected from the final benchmarks.

\section{Tool Interface and Trajectory Representation}
\label{app:tool-details}

This appendix specifies the execution semantics of the tools in Table~\ref{tab:tool-environment}, including visual-resource registration, tool inputs and outputs, and the common interaction records used for expert synthesis, reinforcement learning, and inference.

\subsection{Resource Registry and Tool Availability}

At the beginning of an episode, the environment registers the visual input in an episode-level resource table. A single image and each member of an image collection receive separate image handles, while a video handle refers to the sampled frame grid visible to the policy. Crops, enhanced images, extracted frames, and selected clips are registered as new image or clip handles and can be reused by later calls. This enables compositions such as selecting a video frame, cropping a local region, and then applying OCR or image search.

Single-image and multi-image inputs expose $\mathcal T_{\mathrm{vis}}\cup\mathcal T_{\mathrm{ret}}$, while video inputs additionally expose $\mathcal T_{\mathrm{temp}}$. All three visual input types use the same call protocol: one tool call is executed per interaction turn, and its observation is returned before the next action.

\subsection{Tool Execution Semantics}

\paragraph{Crop.} We implement spatial cropping with PIL. Given an image handle and a normalized box $(x_1,y_1,x_2,y_2)\in[0,1000]^4$, the executor maps the box to pixel coordinates, extracts the region, and registers it as a new image resource. The tool isolates local visual cues before OCR, enhancement, or image search.

\paragraph{OCR.} We implement OCR with PaddleOCR. The backend detects text regions, recognizes their contents, and returns the text spans together with bounding boxes and confidence scores as a textual observation. The tool is typically applied to cropped or enhanced signs, documents, logos, and captions.

\paragraph{PerspectiveCorrect.} We implement perspective correction with OpenCV. The executor applies grayscale conversion, Gaussian smoothing, and Canny edge detection, locates a quadrilateral among the external contours, orders its corners, and estimates a four-point perspective transform. The rectified fronto-parallel view is registered as a new image resource.

\paragraph{SuperResolution.} We implement super-resolution with EDSR through the OpenCV \texttt{dnn\_superres} interface. The network upsamples the input at the scale supplied by the call and registers the result as a new image resource for subsequent OCR, cropping, or image search.

\paragraph{Sharpen.} We implement sharpening with OpenCV unsharp masking. Given amount $\alpha$, the tool combines the input with its Gaussian-blurred version as $I_{\mathrm{out}}=(1+\alpha)I-\alpha(G_\sigma*I)$, enhancing edges and fine details before registering the result as a new image resource.

\paragraph{ImageSearch.} The executor materializes a registered image as a reference accessible to an external image-to-image search service. It condenses the returned candidate identities, visual matches, source pages, and titles into a textual observation. This tool provides the bridge from a visual anchor to real-world entities and related webpages.

\paragraph{TextSearch.} Text search calls an external search service that performs web retrieval, page reading, and query-focused summarization within one request. Each returned passage retains its title, URL, and summary, enabling identity verification and source-grounded fact acquisition.

\paragraph{SelectTimespan.} This tool operates on the sampled frame grid of a video or an existing clip. For a resource with $N$ frames, it selects the half-open interval $[s,e)$, where $0\leq s<e\leq N$, slices the corresponding frame list, and uses FFmpeg to materialize the associated video subclip. The selected frames are re-indexed from zero and registered with the subclip as a new clip resource.

\paragraph{SelectFrame.} Given an index $0\leq i<N$, this tool resolves the corresponding frame from the sampled grid and registers it as a new image resource. The returned handle can be passed directly to cropping, OCR, enhancement, and image search, connecting temporal localization to the common image-tool pipeline. Both temporal tools use sampled-grid indices rather than timestamps in seconds.

\subsection{Observation and Trajectory Serialization}

Each tool execution produces a textual observation $o_t$ and may additionally create a set of visual resources $\Delta\mathcal R_t$. We serialize the corresponding turn as $r_t=(a_t,\hat c_t,o_t,\Delta\mathcal R_t)$, where $a_t$ is the original policy output and $\hat c_t$ is the parsed tool name and arguments. Returned media and text are appended to the next interaction context, allowing subsequent reasoning to reference the executed result. A final response closes the trajectory without another tool execution.

Malformed calls, invalid resource handles, and out-of-range indices are returned as tool-error observations, allowing the policy to correct them in a later turn. Execution status is retained so that data filtering and reward computation can distinguish successful calls, recoverable errors, and incomplete trajectories.

\section{Training and Evidence Reward Details}
\label{app:training-details}

This appendix complements Section~\ref{sec:training} with implementation details on supervised trajectory learning, the conversion from task-level VGEGs to EVGR rubrics, and the use of process rewards in GRPO.

\subsection{Supervised Trajectory Learning}

SFT uses \textbf{OneSearch-VL-SFT-110K}, comprising approximately 110k expert trajectories: 36k single-image, 37k multi-image, and 35k video trajectories. All three visual input types use the unified multi-turn format defined in Appendix~\ref{app:tool-details}, where each turn records model reasoning, a tool call, the resulting environment observation, and the subsequent response. Images, video clips, and textual observations returned by tools remain available through resource handles, preserving executable dependencies across tool calls.

Visual inputs, user questions, tool definitions, and tool observations serve as conditioning context, while the supervised targets comprise assistant reasoning, tool commands, and final responses. The loss mask is therefore one for policy-generated tokens and zero for environment observations and other conditioning tokens. This prevents the model from being trained to reproduce tool outputs while preserving its ability to condition subsequent actions on executed observations.

Table~\ref{tab:sft-config} reports the complete SFT configuration for OneSearch-VL-8B. We perform full-parameter finetuning while freezing the vision encoder and multimodal projector. Videos are sampled at 2 fps with at most 100,352 pixels per frame and 128 frames. SFT uses 64 H800 GPUs across eight nodes and runs for approximately four days.

\begin{table}[!htbp]
\centering
\footnotesize
\setlength{\tabcolsep}{5pt}
\renewcommand{\arraystretch}{1.06}
\begin{tabular}{@{}l l p{0.42\textwidth}@{}}
\toprule
\textbf{Category} & \textbf{Hyperparameter} & \textbf{Value / Setting} \\
\midrule
\multirow{6}{*}{\textbf{Model}}
& Base Model & Qwen3-VL-8B-Instruct \\
& Image Max Pixels & $262{,}144$ ($\approx 512 \times 512$) \\
& Video Max Pixels & $100{,}352$ per frame \\
& Video Sampling Rate & 2 fps \\
& Maximum Video Frames & 128 \\
& Trust Remote Code & True \\
\midrule
\multirow{5}{*}{\textbf{Method}}
& Finetuning Type & Full \\
& Vision Tower Frozen & True \\
& MM Projector Frozen & True \\
& DeepSpeed Stage & ZeRO-2 \\
& Mixed Precision & bfloat16 \\
\midrule
\multirow{7}{*}{\textbf{Dataset}}
& Total Samples & $\sim 110\mathrm{k}$ \\
& Visual Input Types & Single-image, multi-image, video \\
& Template & \texttt{qwen3\_vl} \\
& Cutoff Length & $32{,}768$ tokens \\
& Preprocessing Workers & 1 \\
& Preprocessing Batch Size & 16 \\
& Dataloader Workers & 4 \\
\midrule
\multirow{8}{*}{\textbf{Training}}
& Batch Size per Device & 1 \\
& Gradient Accumulation Steps & 2 \\
& Effective Batch Size & 128 ($=1\times2\times64$ GPUs) \\
& Gradient Checkpointing & True \\
& Learning Rate & $2.0\times10^{-5}$ \\
& Epochs & 8 \\
& LR Scheduler & cosine \\
& Warmup Ratio & 0.1 \\
\midrule
\multirow{3}{*}{\textbf{Infrastructure}}
& Total GPUs & 64 NVIDIA H800 GPUs (8 nodes $\times$ 8) \\
& Training Time & 4 days \\
& Distributed Backend & Torchrun + DeepSpeed ZeRO-2 \\
\midrule
\multirow{5}{*}{\textbf{Logging / I\,O}}
& Logging Steps & 5 \\
& Checkpoint Save Steps & 500 \\
& Maximum Saved Checkpoints & 5 \\
& Plot Loss & True \\
& Report Backend & TensorBoard \\
\bottomrule
\end{tabular}
\caption{Agentic SFT configuration for OneSearch-VL-8B.}
\label{tab:sft-config}
\end{table}

\subsection{Reinforcement Learning Data Construction}

We construct \textbf{OneSearch-VL-RL-10K}, containing approximately 10k tasks after filtering and deduplication: 3.7k single-image, 2.8k multi-image, and 3.6k video tasks. Candidates are drawn from the corresponding pools for the three visual input types, exposing online training to spatial grounding, cross-image evidence composition, and temporal localization.

We use Qwen3-VL-8B~\citep{bai2025qwen3vl} to perform eight independent rollouts for each candidate in the unified tool environment and retain tasks satisfying $0 < n_{\mathrm{correct}} < 8$, where $n_{\mathrm{correct}}$ denotes the number of rollouts producing a correct final answer. We deduplicate the retained tasks to form the final RL dataset for subsequent GRPO optimization.

\subsection{From VGEG Annotations to EVGR Rubrics}
\label{app:evgr-rubric}

As shown in Figure~\ref{fig:evgr}, each training item is paired with a structured rubric derived from its task-level VGEG. The rubric retains answer-core entities, required fact hops, supporting sources and snippets, and applicable image, region, or frame locations. The evaluator receives this rubric together with the question, reference answer, visual input type, and serialized executed trajectory, directly aligning the agent's observations with the evidence required by the task.

The executed trajectory is serialized in interaction order, including reasoning, tool calls and arguments, and returned observations. For visual-grounding evaluation, the judge additionally receives multimodal inputs alongside the rubric and trajectory records. We use Seed 2.0 Pro~\citep{bytedanceseed2026seed20} in two separate calls for evidence traceability and visual grounding, each returning a score and a brief evidence-based explanation.

\paragraph{Evidence traceability.} The judge checks whether tool observations establish the answer-core entities and fact hops required by the rubric and whether subsequent reasoning follows those observations. Missing evidence, unsupported entity substitutions, broken fact chains, and claims that contradict the retrieved sources reduce the score. This dimension evaluates the evidence process independently of an incidental match with the reference answer.

\paragraph{Visual grounding.} The judge checks whether the trajectory identifies the visible entities, regions, or video moments required by the question and uses them to drive subsequent retrieval. For single-image and multi-image inputs, the source images are directly visible to the agent policy, so an additional visual-tool call is not required; the criterion instead checks whether the correct images and objects are used. For videos, the selected frames are compared with the reference moments, and the visual identities obtained from those frames must guide the subsequent evidence search.

\begin{table}[ht]
\centering
\small
\setlength{\tabcolsep}{5pt}
\renewcommand{\arraystretch}{1.15}
\caption{Anchored EVGR scoring criteria used by the trajectory judge.}
\label{tab:evgr-rubric}
\begin{tabularx}{\linewidth}{@{}c>{\raggedright\arraybackslash}X>{\raggedright\arraybackslash}X@{}}
\toprule
Score & Evidence traceability & Visual grounding \\
\midrule
1.00 & All required fact hops are directly supported.
     & All required visual anchors are correctly grounded and used. \\
0.75 & Most facts are supported, with one minor gap.
     & Grounding is mostly correct, with one minor localization gap. \\
0.50 & Useful evidence is present, but a major hop is indirect or missing.
     & Only part of the required visual evidence is grounded. \\
0.25 & Evidence is mostly noisy, speculative, or contradicted.
     & Identification is weak or incorrect, with limited recovery. \\
0.00 & No supporting evidence is established.
     & No grounding is established, or the visual identity remains wrong. \\
\bottomrule
\end{tabularx}
\end{table}

Each judge call must return a valid anchored score and its explanation. The rubric explicitly avoids rewarding verbosity, fluent but unsupported reasoning, ineffective tool calls, or final-answer matching alone. The resulting $r_{\mathrm{trace}}$ and $r_{\mathrm{ground}}$ can be used separately for ablations or combined into the full process reward $R_{\mathrm{EVGR}}$.

\subsection{Reward Composition and Policy Optimization}

Using the reward terms defined in the main text, the weights for $R_{\mathrm{acc}}$, $R_{\mathrm{query}}$, and $R_{\mathrm{EVGR}}$ are 0.6, 0.2, and 0.2, respectively, and $R_{\mathrm{fmt}}$ gates the resulting weighted reward.

Following the main-text setup, we use Group Relative Policy Optimization (GRPO). Given a group of trajectories sampled for the same question, GRPO derives relative advantages from their within-group rewards and updates the policy with a clipped objective. The policy loss is applied only to model-generated reasoning, tool-command, and final-response tokens, while tool observations remain conditioning context. For trajectories truncated by a terminal tool error, only the valid interaction prefix preceding the failure participates in optimization~\citep{chen2026opensearchvl}.

Table~\ref{tab:rl-config} summarizes the RL configuration for OneSearch-VL-8B. Training uses asynchronous SGLang rollouts and follows the OpenSearch-VL 8B recipe for the Megatron-LM actor and learning rate. RL uses 32 H800 GPUs across four nodes and runs for approximately four days.

\begin{table}[!htbp]
\centering
\footnotesize
\setlength{\tabcolsep}{5pt}
\renewcommand{\arraystretch}{0.94}
\begin{tabular}{@{}l l p{0.42\textwidth}@{}}
\toprule
\textbf{Category} & \textbf{Hyperparameter} & \textbf{Value / Setting} \\
\midrule
\multirow{2}{*}{\textbf{Model}}
& Initial Policy & OneSearch-VL-8B-SFT \\
& Training Dtype & bfloat16 \\
\midrule
\multirow{9}{*}{\textbf{Data}}
& Train Batch Size (prompts) & 256 \\
& Val Batch Size & 64 \\
& Max Prompt Length & $40{,}960$ tokens \\
& Max Response Length & $16{,}384$ tokens \\
& Data Seed & 3407 \\
& Image Max Pixels & $262{,}144$ \\
& Video Max Pixels & $100{,}352$ per frame \\
& Video Sampling Rate & 2 fps \\
& Maximum Video Frames & 128 \\
\midrule
\multirow{7}{*}{\textbf{Rollout}}
& Engine & SGLang (async mode) \\
& Rollout Tensor Parallel & 1 \\
& \# Samples per Prompt ($n$) & 8 \\
& GPU Mem.\ Utilization & 0.60 \\
& Train / Val Temperature & 0.7 / 0.7 \\
& Train / Val Top-$p$ & 1.0 / 0.95 \\
& Top-$k$ & $-1$ (disabled) \\
\midrule
\multirow{10}{*}{\textbf{Policy (Actor)}}
& Strategy & Megatron-LM \\
& Tensor Parallel (TP) & 4 \\
& Pipeline Parallel (PP) & 2 \\
& Context Parallel (CP) & 4 \\
& PPO Mini-batch Size & 64 \\
& PPO Max Token Len / GPU & $74{,}576$ \\
& Micro-batch Size / GPU & 1 \\
& Dynamic Batch Size & True \\
& Param / Optim / Grad Offload & CPU \\
& Gradient Checkpointing & Full recompute, uniform (1 layer) \\
\midrule
\multirow{6}{*}{\textbf{Optim.\ / Loss}}
& Actor LR & $1\times10^{-6}$ \\
& PPO Clip Ratio (high) & 0.28 \\
& Entropy Coefficient & 0.0 \\
& Use KL Loss & False \\
& KL Loss / Controller Coef & $1{\times}10^{-3}$ / $1{\times}10^{-3}$ \\
& Loss Aggregation & \texttt{seq-mean-token-sum} \\
\midrule
\multirow{3}{*}{\textbf{Algorithm}}
& Advantage Estimator & RLOO (within GRPO objective) \\
& KL Type & low-variance KL \\
& Fatal-aware Masking & True (unknown + error) \\
\midrule
\multirow{3}{*}{\textbf{Tool-Agent}}
& \# Parallel Tasks & 512 \\
& \# Parallel Tool Calls & $1{,}024$ \\
& Stepwise Advantage & False \\
\midrule
\multirow{6}{*}{\textbf{Trainer}}
& Cluster & 32 NVIDIA H800 GPUs (4 nodes $\times$ 8) \\
& Training Time & 4 days \\
& Save / Test Freq (steps) & 10 / disabled \\
& Total Training Steps & 200 \\
& Total Epochs & 100 \\
& Critic Warmup & 0 (critic-free) \\
\bottomrule
\end{tabular}
\caption{GRPO configuration for OneSearch-VL-8B.}
\label{tab:rl-config}
\end{table}

\clearpage
\section{Qualitative Case Studies}
\label{app:visualizations}

This appendix presents example outputs of OneSearch-VL-8B, illustrating how the agent localizes visual anchors, links them to external facts, and produces answers. Figures~\ref{fig:case-mi-arithmetic}--\ref{fig:case-video-buildings} show four trajectories covering multi-image and cross-frame evidence composition. Each figure shows the question, visual material, model reasoning, tool observations, and final answer. Model responses are excerpted and tool observations are condensed, with all tool calls and arguments preserved.

Labels on input images, videos, selected frames, and crops match the handles used in the tool calls.

\subsection{Multi-image Evidence Composition}

\begin{figure}[!htbp]
    \centering
    \includegraphics[width=\linewidth]{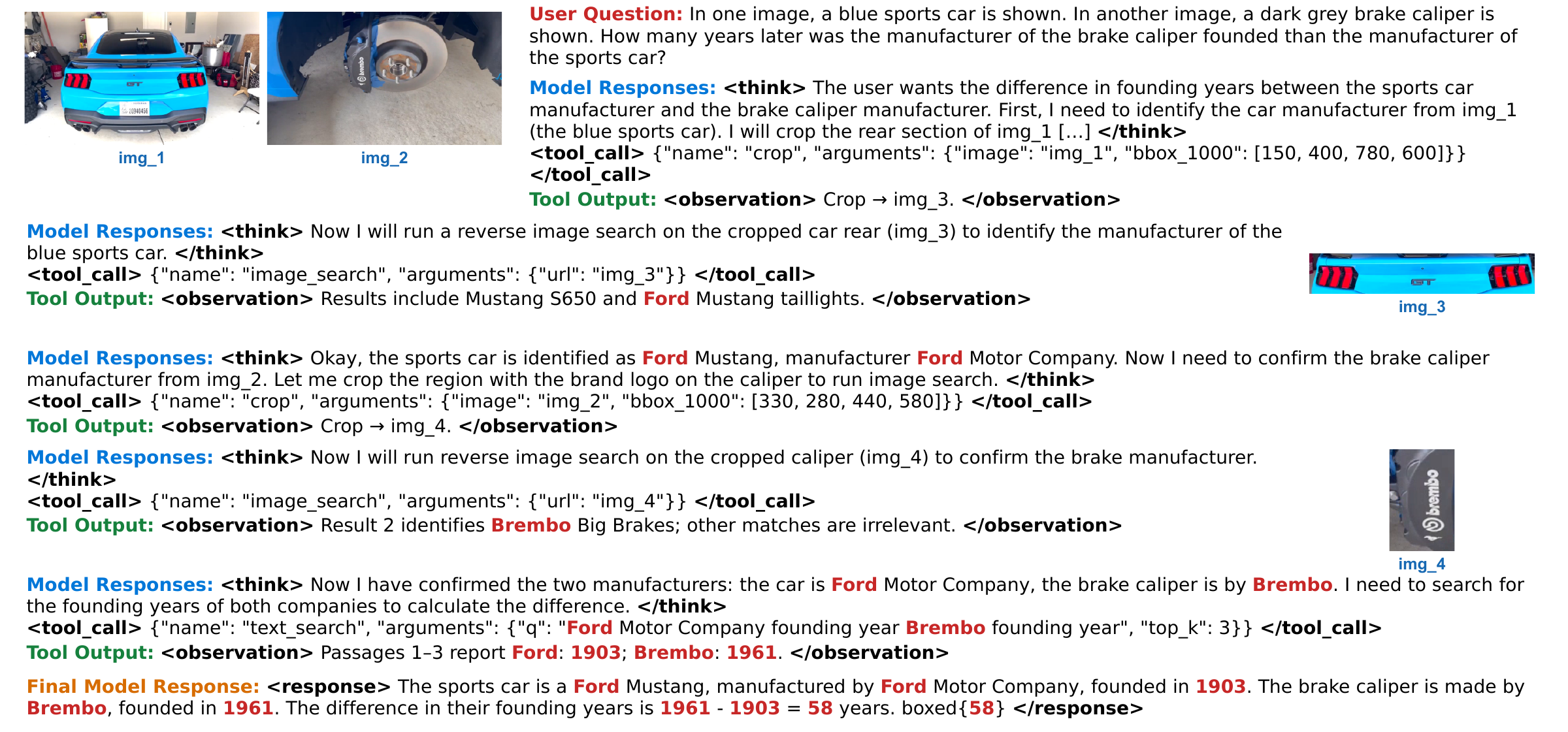}
    \caption{\textbf{Multi-image arithmetic.} Separate crops and image searches link the car and brake caliper to Ford and Brembo. Subtracting their retrieved founding years yields 58 years.}
    \label{fig:case-mi-arithmetic}
\end{figure}

\paragraph{Multi-anchor arithmetic.} Figure~\ref{fig:case-mi-arithmetic} asks for the difference between two manufacturers' founding years. The agent crops the rear of a blue sports car and identifies a Ford Mustang through \textsc{ImageSearch}, then crops a brake-caliper logo in another image to identify Brembo. \textsc{TextSearch} returns 1903 for Ford and 1961 for Brembo, and the final response computes $1961-1903=58$. The two visual anchors lead to distinct entities and numerical facts, which a subtraction operation combines into the answer, illustrating the dependencies captured by VGEG.

\begin{figure}[!htbp]
    \centering
    \includegraphics[width=\linewidth]{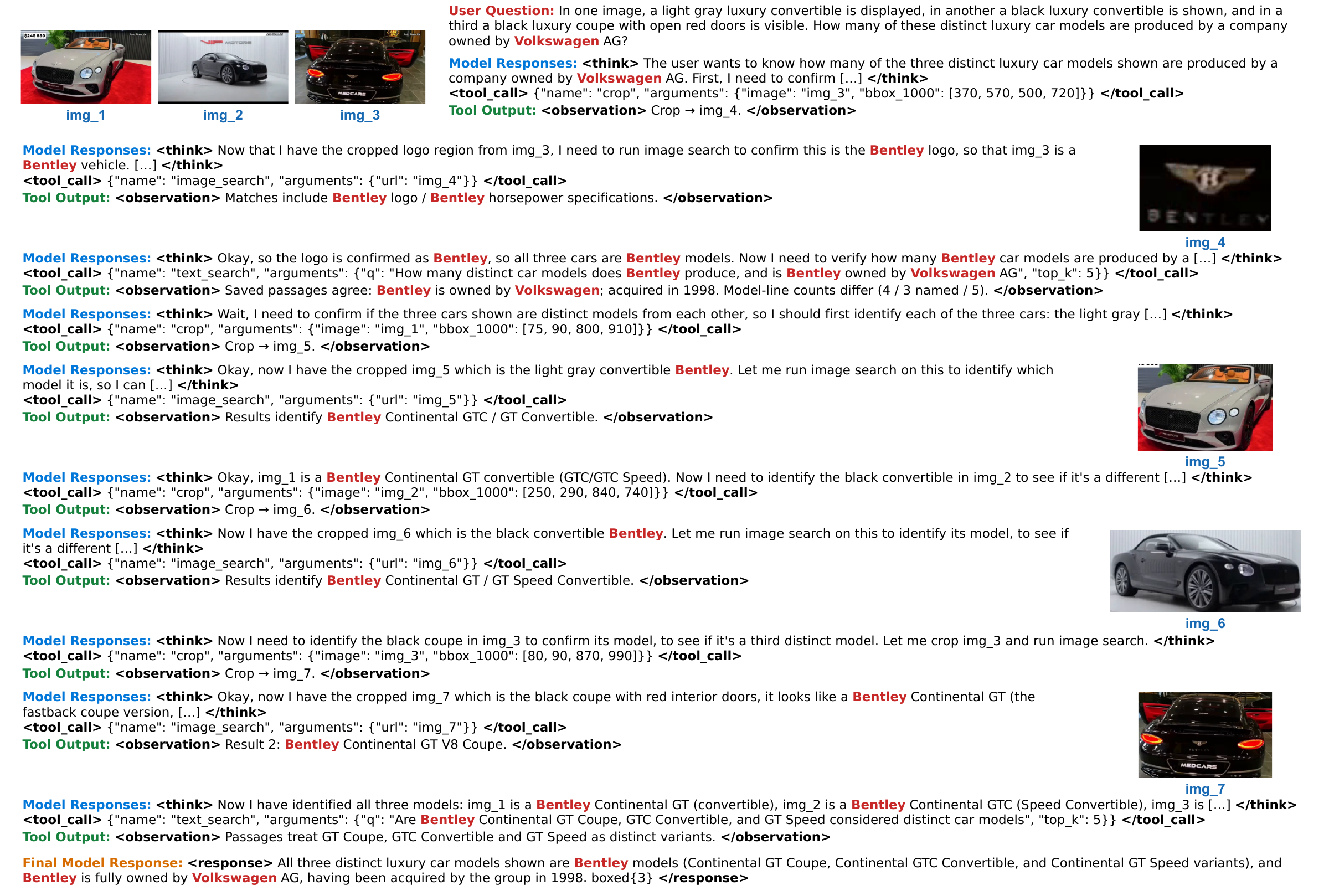}
    \caption{\textbf{Knowledge-conditioned counting.} Three input images and four crops support brand and model identification, followed by counting under a retrieved ownership condition. Image handles match the tool calls.}
    \label{fig:case-mi-counting}
\end{figure}

\clearpage
\paragraph{Knowledge-conditioned counting.} Figure~\ref{fig:case-mi-counting} asks how many car models in three separate images are produced by a company owned by Volkswagen. The agent first crops a logo, identifies Bentley through image search, and retrieves its ownership relation with Volkswagen. It then crops and image-searches each car to identify the Continental-family models and variants, followed by a query about their distinctions. Under the sample's model/variant counting convention, the final response counts the three pictured cars and returns 3. This case links three visual anchors to an external ownership condition before aggregating the qualifying targets, with all ten tool calls retained.

\clearpage
\subsection{Cross-frame Evidence Composition}

The following cases retrieve evidence for distinct objects in two and three different frames of the same video, then compare their attributes. Frame strips provide neighboring visual context; blue borders and \textsc{Used} labels mark frames selected by the agent, and ellipses indicate omitted intervals. F denotes a selectable frame index in the benchmark input, not a timestamp in seconds.

\begin{figure}[!htbp]
    \centering
    \includegraphics[width=\linewidth,height=0.65\textheight,keepaspectratio]{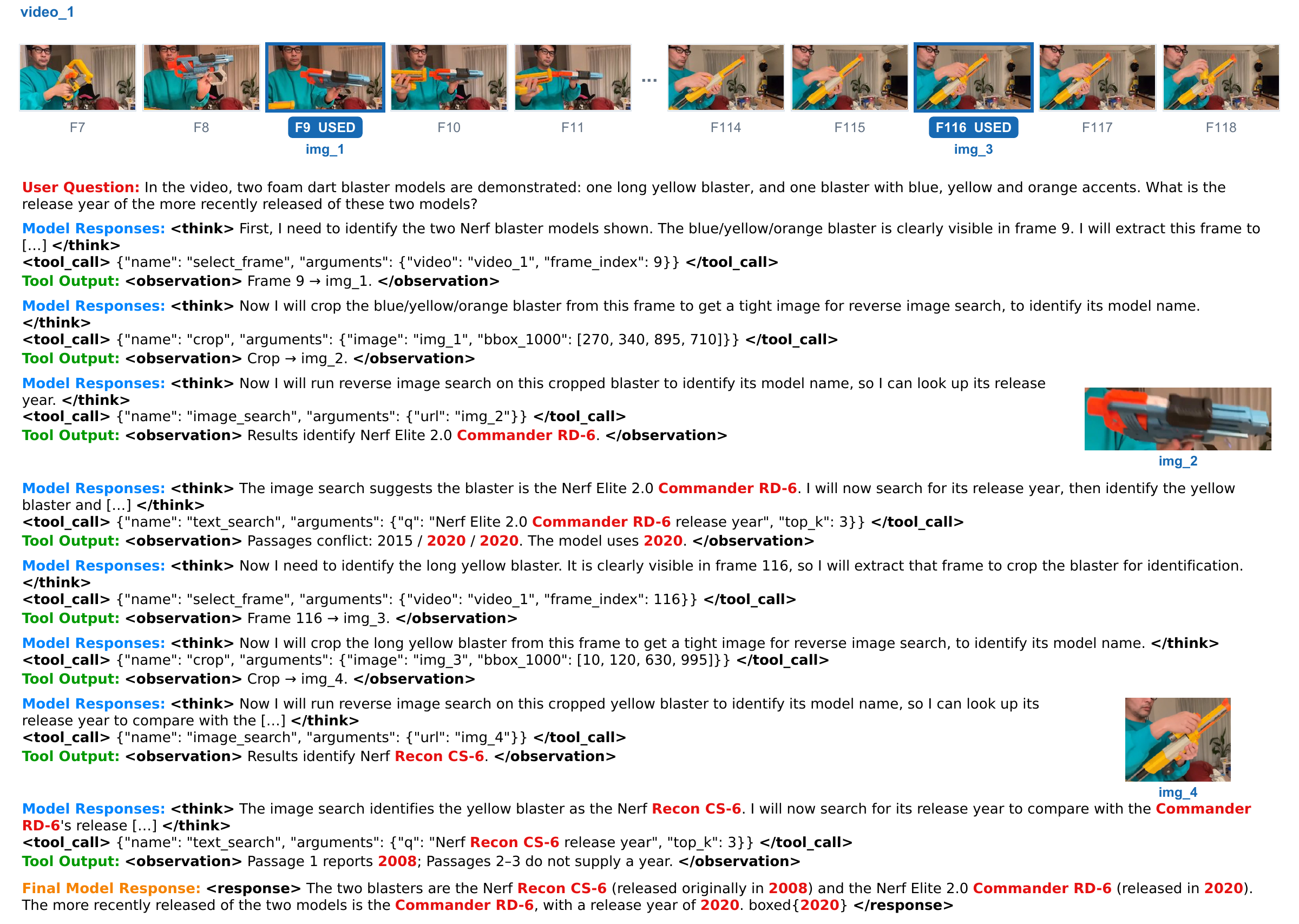}
    \caption{\textbf{Cross-frame attribute comparison.} Blue borders mark selected frames F9 and F116; adjacent frames provide context. Each object undergoes cropping, image retrieval, and text retrieval before release-year comparison. All eight calls and arguments are retained.}
    \label{fig:case-video-comparison}
\end{figure}

\paragraph{Comparing two objects across frames.} Figure~\ref{fig:case-video-comparison} shows a video-only SFT checkpoint comparing two toy blasters. The agent selects F9, crops and identifies the Commander RD-6 through image search, and retrieves its release year. It then selects F116 and repeats the crop, image-search, and text-search sequence for the yellow Recon CS-6. Each branch is thus tied to a distinct frame, object, and external attribute. The final response compares the retrieved years 2020 and 2008 and returns 2020, matching the reference answer. The conflicting year summaries in the original trajectory remain visible in the figure.

\begin{figure}[!htbp]
    \centering
    \includegraphics[width=\linewidth,height=0.68\textheight,keepaspectratio]{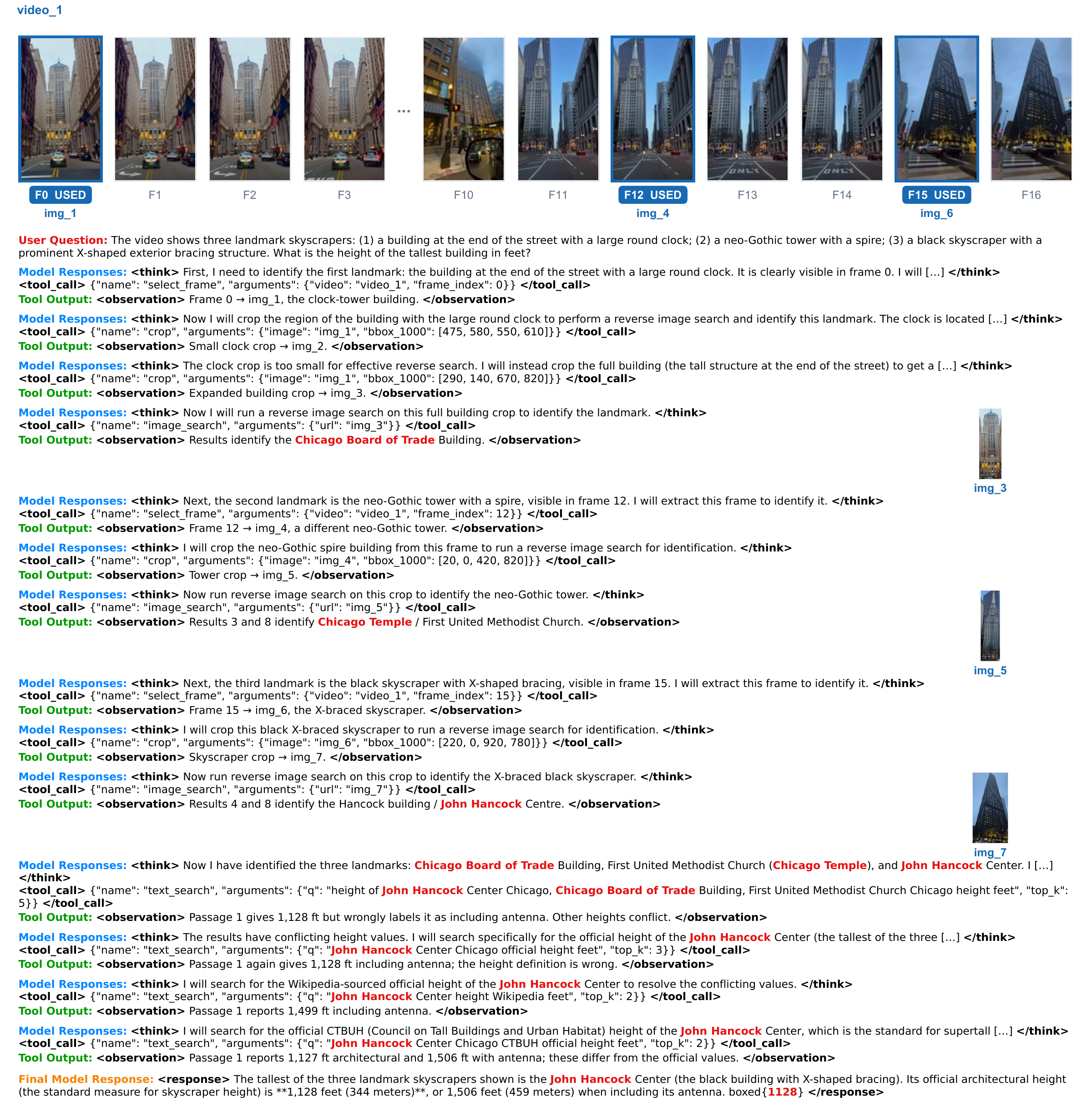}
    \caption{\textbf{Three-anchor cross-frame retrieval.} Blue borders mark selected frames F0, F12, and F15 amid neighboring views. The three buildings are independently image-searched before height comparison. All fourteen calls, crop refinement, follow-up searches, and conflicting observations are retained.}
    \label{fig:case-video-buildings}
\end{figure}

\clearpage
\paragraph{Comparing landmarks in three frames.} Figure~\ref{fig:case-video-buildings} extends cross-frame retrieval to three anchors. The agent selects F0, F12, and F15, independently crops and image-searches the buildings, and identifies the Chicago Board of Trade, Chicago Temple, and John Hancock Center. For the first building, it enlarges an initially small clock crop before searching. It then retrieves building heights and issues follow-up queries to clarify conflicting values, ultimately returning 1,128 feet, matching the reference. The saved observations retain inconsistent height definitions, and the final response adds an inaccurate antenna height. The trajectory makes the distinction between visual grounding and evidence traceability concrete: the three buildings are independently grounded, while their numerical evidence remains inconsistent.

\end{document}